\documentclass[10pt,twocolumn,letterpaper]{article}

\usepackage[pagenumbers]{cvpr}              %

\usepackage{graphicx}
\usepackage{amsmath}
\usepackage{amssymb}
\usepackage{booktabs}
\usepackage{bbm}
\usepackage{multirow}
\usepackage{multicol}
\usepackage{caption}
\usepackage{array}
\usepackage{makecell}

\usepackage[dvipsnames]{xcolor}
\usepackage{textcomp}
\usepackage{arydshln}
\usepackage{cuted}

\usepackage{pifont}%

\newcommand*\xbar[1]{%
  \hbox{%
    \vbox{%
      \hrule height 0.5pt %
      \kern0.3ex%
      \hbox{%
        \kern-0.2em%
        \ensuremath{#1}%
        \kern-0.05em%
      }%
    }%
  }%
} 

\usepackage{enumitem}
\setlist{itemsep=0.1cm,topsep=0.1cm,parsep=0.1cm,partopsep=0pt,leftmargin=0.4cm}

\usepackage[pagebackref,breaklinks,colorlinks]{hyperref}

\usepackage[capitalize]{cleveref}
\crefname{section}{Sec.}{Secs.}
\Crefname{section}{Section}{Sections}
\Crefname{table}{Table}{Tables}
\crefname{table}{Tab.}{Tabs.}

\def\cvprPaperID{8411} %
\def\confName{CVPR}
\def\confYear{2023}

\begin{document}

\title{A Light Touch Approach to Teaching Transformers Multi-view Geometry}

\author{Yash Bhalgat \quad Jo\~{a}o F.\ Henriques \quad Andrew Zisserman
\\
\vspace{10\baselineskip}
Visual Geometry Group \\
University of Oxford \\
{\tt\small \{yashsb,joao,az\}@robots.ox.ac.uk}
}
\maketitle

\begin{abstract}
Transformers are powerful visual learners, in large part due to their conspicuous lack of manually-specified priors. This flexibility can be problematic in tasks that involve multiple-view geometry, due to the near-infinite possible variations in 3D shapes and viewpoints (requiring flexibility), and the precise nature of projective geometry (obeying rigid laws).
To resolve this conundrum, we propose a ``light touch'' approach, guiding visual Transformers to learn multiple-view geometry but allowing them to break free when needed.
We achieve this by using epipolar lines to guide the Transformer's cross-attention maps during training, penalizing attention values outside the epipolar lines and encouraging higher attention along these lines since they contain geometrically plausible matches. 
Unlike previous methods, our proposal does not require any camera pose information at test-time.
We focus on pose-invariant object instance retrieval, where standard Transformer networks struggle, due to the large differences in viewpoint between query and retrieved images.
Experimentally, our method
outperforms state-of-the-art approaches at object retrieval, without needing pose information at test-time.
\end{abstract}

\section{Introduction}
\label{sec:intro}
Recent advances in computer vision have been characterized by using increasingly generic models fitted with large amounts of data, with attention-based models (e.g. Transformers) at one extreme \cite{ViT,DeTR,fedus2021switch,liu2021swin,DINO,jaegle2021perceiver}.
There are many such recent examples, where shedding priors in favour of learning from more data has proven to be a successful strategy, from image classification \cite{ViT,yuan2021tokens,TnT,DINO,xcit},
action recognition \cite{neimark2021video,girdhar2019video,plizzari2021spatial,fan2021multiscale,bertasius2021space}, to text-image matching 
\cite{CLIP,ALIGN,Su2020VL-BERT,lu2019vilbert} and 3D recognition \cite{zhao2021point,lin2021end}.
One area where this strategy has proven more difficult to apply is solving tasks that involve reasoning about multiple-view geometry, such as object retrieval -- i.e.\ finding all instances of an object in a database given a single query image. 
This has applications in image search \cite{nie2007web, van2009visual, jing2008pagerank, 5540092, zhou2010spatial},
including identifying landmarks from images \cite{noh2017large,weyand2020GLDv2,radenovic2018revisiting}, recognizing artworks in images \cite{ufer2021large}, 
retrieving relevant product images in e-commerce databases \cite{oh2016deep,Le_2020_ECCV} or retrieving specific objects from a scene \cite{Arandjelovic11,johnson2015image, rabinovich2007objects, ma2020spatial}.

\begin{figure}
    \centering
    \includegraphics[width=\linewidth]{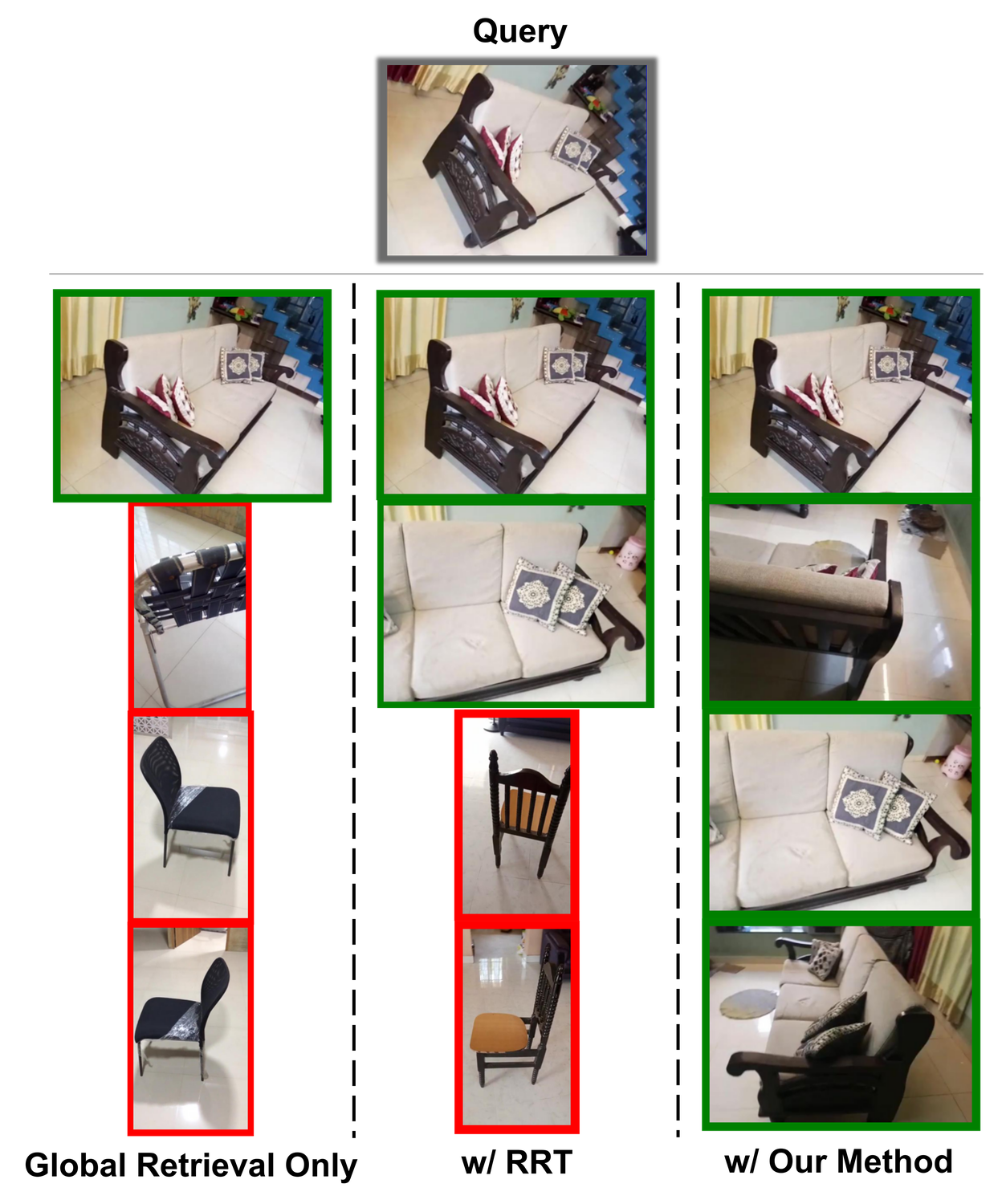}
    \caption{Top-$4$ retrieved images 
    with (1) global retrieval (left column),
    (2) Reranking Transformer (RRT) \cite{rrt} (middle), and
    (3) RRT trained with
    our proposed Epipolar Loss (right column).
    Correct retrievals are \textbf{\textcolor{OliveGreen}{green}}, incorrect ones are \textbf{\textcolor{red}{red}}.
    The Epipolar Loss imbues RRT with an implicit geometric understanding, allowing it to match images from extremely diverse viewpoints.
    }
    \label{fig:my_label}
\end{figure}

The main challenges in object retrieval include overcoming variations in viewpoint and scale. The difficulty in viewpoint-invariant object retrieval can be partially explained by the fact that it requires disambiguating similar objects by small differences in their unique details, which can have a smaller impact on an image than a large variation in viewpoint.
For this reason, several works have emphasized geometric priors in deep networks that deal with multiple-view geometry \cite{CamConv,Yifan_2022_CVPR}.
It is natural to ask whether these priors are too restrictive, and harm a network's ability to model the data when it deviates from the geometric assumptions.
As a step in this direction, we explore how to ``guide'' attention-based networks with soft guardrails that encourage them to respect multi-view geometry, without constraining them with any rigid mechanism to do so.

\begin{figure*}
    \centering
    \includegraphics[width=\textwidth]{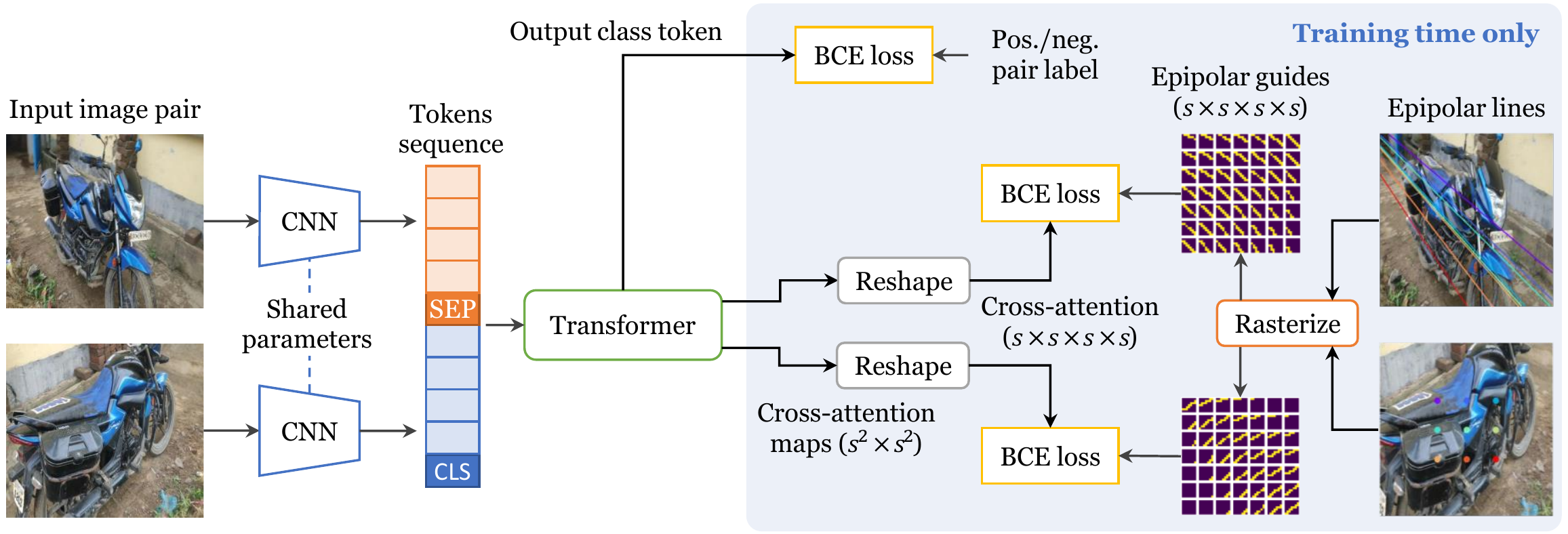}
    \caption{Overview of the proposed method.
    Features from two candidate images are extracted with a Convolutional Neural Network, and concatenated into a sequence of tokens for a Transformer. They are separated by a learned $\langle\text{SEP}\rangle$ token and end with a $\langle\text{CLS}\rangle$ token.
    The model is trained with a Binary Cross Entropy (BCE) loss to predict whether the two images match.
    During training, epipolar lines relating the two views (obtained with ground truth camera information) are rasterized into 4D tensors. These ``epipolar guides'' denote matches that are geometrically plausible given the viewpoints, and are used to train the Transformer's cross-attention maps using BCE losses.
    }
    \label{fig:mainfig}
    \vspace*{-10pt}
\end{figure*}

In this work, we focus on post-retrieval reranking methods, wherein an initial ranking is obtained using global (image-level) representations and then local (region- or patch-level) representations are used to \textit{rerank} the top-ranked images either with the classic Geometric Verification \cite{Philbin07}, or by directly predicting similarity scores of image pairs using a trained deep network \cite{rrt,hausler2021patch}. Reranking can be easily combined with any other retrieval method while significantly boosting the precision of the underlying retrieval algorithm. Recently, PatchNetVLAD \cite{hausler2021patch}, DELG \cite{cao2020unifying}, and Reranking Transformers \cite{rrt} have shown that learned reranking can achieve state-of-the-art performance on object retrieval. We show that the performance of such reranking methods can be further improved by \textit{implicitly} inducing geometric knowledge, specifically the epipolar relations between two images arising from relative pose, into the underlying image similarity computation. 

This raises the question of whether multiple view relations should be incorporated into the two view architecture \textit{explicitly} rather than \textit{implicitly}. In the explicit case, the epipolar relations between the two images are supplied as inputs. For example, this is the approach taken in the Epipolar Transformers architecture~\cite{he2020epipolar} where candidate correspondences are explicitly sampled along the epipolar line, and in~\cite{Yifan_2022_CVPR} where pixels are tagged with their epipolar planes using a Perceiver IO architecture \cite{jaegle2021perceiver}. 
The disadvantage of the explicit approach is that epipolar geometry must be supplied at inference time, requiring a separate process for its computation, and being problematic
when images are not of the same object (as the epipolar geometry is then not defined). In contrast, in the implicit approach the epipolar geometry is only required at training time and is applied as a loss to encourage the model to learn to (implicitly) take advantage of epipolar constraints when determining a match.

We bring the following three contributions in this work:
First, we propose a simple but effective {\em Epipolar Loss} to induce epipolar constraints into the cross-attention layer(s) of transformer-based reranking models. We only need the relative pose (or epipolar geometry) information during training to provide the epipolar constraint. Once trained, the reranking model develops an implicit understanding of the relative geometry between any given image pair and can effectively match images containing an object instance from very diverse viewpoints \emph{without} any additional input. Second, we set up an object retrieval benchmark on top of the CO3Dv2~\cite{co3d} dataset which contains ground-truth camera poses and provide a comprehensive evaluation of the proposed method, including a comparison between implicit and explicit incorporation of epipolar constraints.
The benchmark configuration is detailed in Sec.~\ref{sec:co3d-retrieve}.
Third, we evaluate on the Stanford Online 
Products~\cite{oh2016deep} dataset using both zero-shot and fine-tuning, outperforming previous methods on this standard object instance retrieval benchmark.

\section{Related Work}
\label{sec:relwork}

\noindent\textbf{Computing epipolar geometry.} 
Estimating epipolar geometry given an image pair is a fairly broad problem, well-studied in multi-view geometry and computer vision \cite{Hartley04c}. Classic techniques involve predicting interest points and their descriptors \cite{Mikolajczyk05,lowe1999object,Arandjelovic12,rosten2006machine,rublee2011orb} in the images and finding point correspondences to estimate the relative geometry \cite{longuet1981computer,nister2004efficient}. Several learning based methods have been proposed to provide improved interest point detection and features, e.g.\ R2D2 \cite{revaud2019r2d2} SuperPoint \cite{detone2018superpoint}, LIFT \cite{yi2016lift} and MagicPoint \cite{detone2017toward}. These features along with learning based local matching methods~\cite{Wiles21,sun2021loftr,sarlin2020superglue} and robust optimization methods \cite{barath2020magsac++,brachmann2017dsac,fischler1981random} form a powerful toolbox for relative geometry estimation. We use a combination of LoFTR \cite{sun2021loftr} and MAGSAC++ \cite{barath2020magsac++} to generate pseudo-geometry information in one of our compared methods.

\noindent\textbf{Incorporating epipolar geometry in Deep Learning.} 
Recently, many works have proposed incorporating geometric priors into deep networks to deal with problems requiring multi-view understanding, such as 3D pose estimation \cite{he2020epipolar,yu2021pcls,rhodin2018unsupervised}, 3D reconstruction \cite{Yifan_2022_CVPR,tulsiani2017multi} or depth estimation \cite{prasad2018epipolar}. Most of these approaches incorporate the epipolar geometry explicitly, e.g.\ Epipolar Transformers \cite{he2020epipolar} compute 3D-aware features for a point by aggregating features sampled on the corresponding epipolar line, which are shown to improve multi-view 3D human-pose estimation. 
\cite{Yifan_2022_CVPR}, another explicit method, proposed a few ways of featurizing multi-view geometry by encoding camera parameters or epipolar plane parameters and using them to provide geometric priors at the input-level. The epipolar plane encoding is also studied in this paper in the context of reranking transformers.
Works such as \cite{rhodin2018unsupervised,tulsiani2018multi} propose implicitly incorporating geometric priors using multi-view consistency. Our work also falls in the implicit category, where we use epipolar constraints as a loss function applied to cross-attention maps to induce geometric understanding.

\noindent\textbf{Image representations for retrieval.}
Traditionally, hand-crafted descriptors such as SIFT~\cite{lowe1999object}, RootSIFT~\cite{Arandjelovic12} and BoVW~\cite{Philbin07} were widely used for object retrieval. However, learned image-level (global) and region-level (local) representations \cite{arandjelovic2016netvlad,babenko2014neural,gordo2016deep,detone2018superpoint,yi2016lift} have shown to surpass the performance of engineered features on large-scale datasets. Local learned representations can also be simply extracted as feature volumes from convolution neural networks or transformer backbones. Global representations are obtained by a combination of (1) downsampling/pooling operations inside a deep network, (2) learned clustering-based pooling operations \cite{arandjelovic2016netvlad} and/or specialized pooling operations such as R-MAC \cite{tolias2015particular}. Hybrid approaches that combine global and local features have also recently been proposed \cite{hausler2021patch,cao2020unifying}.

\noindent \textbf{Post-retrieval reranking}. 
Early reranking methods, such as~\cite{Philbin07,Jegou08}, used Geometric Verification (GV) with local features to compute geometric consistency between the query and reference images. This improved the precision of the top-ranked retrievals. Query Expansion (QE) was used to improve the recall. Popular QE variants such as average-QE and $\alpha$-QE compute an updated query descriptor from the global descriptors of the top retrieved images~\cite{Chum07b,chum2011total,tolias2014visual} and use it to retrieve a new set of top-ranked images. GV can be combined with many deep learning based retrieval methods used today, e.g.\ \cite{cao2020unifying} uses RANSAC based GV on local features from its backbone model. Since RANSAC-based GV can be prohibitively slow for practical applications, \cite{hausler2021patch} proposes a rapid spatial scoring technique as an efficient alternative. Recently, transformer based methods \cite{el2021training,rrt} have been introduced for retrieval and reranking. Our work builds on top of Reranking Transformers \cite{rrt}.

\noindent \textbf{Retrieval with 3D information.}\ Recently, methods using 3D data \cite{uy2018pointnetvlad,liu2019lpd}, structural cues \cite{oertel2020augmenting} or view synthesis \cite{taira2018inloc} have been proposed. Our goal in this work is to build image representations that capture 3D priors and can be used to retrieve images with large variations in pose or scale.

\section{Method}
\label{sec:method}
We describe two variants of our method that \textit{implicitly} or \textit{explicitly} encourage Transformers to use geometric constraints in their predictions.
Our work is built on top of Reranking Transformers (RRT)~\cite{rrt}, a state-of-the-art approach for object retrieval with reranking. The explicit version, inspired by recent work~\cite{Yifan_2022_CVPR}, serves both as a baseline and as a contrast to our proposed implicit approach. We first provide a brief review of RRTs for the reader (Sec.~\ref{sec:rrt}) and then describe our proposed Epipolar Loss (Sec.~\ref{sec:epiloss}), as well as Epipolar Positional Encodings (Sec.~\ref{sec:epiposenc}). The implementation details 
are given in Sec.~\ref{sec:impl}.

\subsection{Review of Reranking Transformers}
\label{sec:rrt}
Post-retrieval reranking is a popular technique used to boost the precision of object retrieval methods, wherein an initial ranking is obtained using global (image-level) descriptors
and then local (region-level) descriptors along with the global ones are used to \textit{rerank} the top-ranked images. In~\cite{rrt}, each image ($\mathcal{I}$) is processed through a ResNet-50 \cite{ResNet} model to extract local features from the last convolution layer, with size $s\times s\times c$ ($s=7$, $c=2048$). Each of the $s^2$ local feature vectors is linearly projected from size $c$ to a smaller size $m=128$. Let these be denoted by $\textbf{x}^l\in\mathcal{R}^{s^2\times m}$ and their 2D positions in the feature volume by $\textbf{p}_{i}\in\mathcal{R}^2$. The global features, computed as the mean of the local features, are used for initial ranking. Then, a light-weight transformer model (4 self-attention heads, 6 layers) is used to rerank these top predictions. With $\mathcal{I}$ as the query and $\mathcal{\bar{I}}$ as a reference image from the top predictions,
as well as class $\langle\text{CLS}\rangle$ and separator $\langle\text{SEP}\rangle$ tokens (consisting of learnable embeddings),
the input to the transformer model is constructed as the concatenation of tokens:
\vspace{-3pt}
\begin{align*}
X(\mathcal{I},\mathcal{\bar{I}}) = [\langle\text{CLS}\rangle, f(\textbf{x}^{l}_{1}), \dots f(\textbf{x}^{l}_{s^2}), \\
\langle\text{SEP}\rangle, \bar{f}(\bar{\textbf{x}}^{l}_{1}), \dots \bar{f}(\bar{\textbf{x}}^{l}_{s^2})]
\end{align*}
where $f(\textbf{x}^{l}_{i})=\textbf{x}^{l}_{i}+\psi(\textbf{p}_{i})+\beta$, $\bar{f}(\bar{\textbf{x}}^{l}_{i})=\bar{\textbf{x}}^{l}_{i}+\psi(\bar{\textbf{p}}_{i})+\bar\beta$, $\psi(.)$ is the frequency position encoding \cite{vaswani2017attention} and $\beta, \bar\beta$ are
learnable embeddings that
differentiate descriptors of $\mathcal{I},\mathcal{\bar{I}}$. Sec.~\ref{sec:impl} provides training details for the reranking model.

\vspace{-1pt}

\subsection{Review of Epipolar Geometry}
Epipolar geometry limits the possible image correspondences for projections of an observed 3D point 
from different viewpoints.
A central concept is the epipolar line. Consider a 2D point $\bf x$ in one image. 
It may correspond to an infinity of 3D points -- one for each possible depth -- which lie on a 3D line that extends from the camera center and passes through $\bf x$ in the image plane.
This 3D line, when projected into a \emph{second} image captured from another viewpoint, is an epipolar line of $\bf x$.
This mapping from a point in one image to its epipolar line in another image can be seen in Fig.~\ref{fig:mainfig} (right).
Epipolar geometry can be used to effectively constrain matches across viewpoints: starting from a point in one image, it can only match points in another image that lie along its epipolar line.
Epipolar geometry can be computed directly from two images,  either from their relative pose or from correspondences, without requiring any information about depth or 3D geometry of the observed scene. Mathematically it is represented by a $3 \times 3$ fundamental matrix. For a more detailed exposition, please refer to~\cite{Hartley04c}.

\subsection{Epipolar Loss}
\label{sec:epiloss}

Given the feature volumes $\textbf{x}^{l},\bar{\textbf{x}}^{l}\in\mathcal{R}^{s^2\times m}$ as input tokens (in addition to $\langle \text{CLS} \rangle$, $\langle \text{SEP} \rangle$), let $\textbf{y}_{L-1},\bar{\textbf{y}}_{L-1}$ denote the corresponding inputs to the last 
transformer layer in the RRT model. The \textit{raw} cross-attention between these outputs can be computed as $A^{12}=Q\xbar{K}^T$ and $A^{21}=\xbar{Q}K^T$, where $W_Q,W_K$ are query and key projection matrices and $Q=W_Q\textbf{y}_{L-1}$, $K=W_K\textbf{y}_{L-1}$, $\xbar{Q}=W_Q\bar{\textbf{y}}_{L-1}$, $\xbar{K}=W_K\bar{\textbf{y}}_{L-1}$.

Next, given the epipolar geometry between the input images, for every location $i\in\{1,\dots,s^2\}$ in $\textbf{x}^{l}$, we can find the set of locations $\bar{e}_i$ in $\bar{\textbf{x}}^{l}$ that lie on the corresponding epipolar line. Similarly, for each location $i\in\{1,\dots,s^2\}$ in $\bar{\textbf{x}}^{l}$, we can find the corresponding set of locations $e_i$ in $\textbf{x}^{l}$. We want to encourage the network, for a given position in the first volume, to only \textit{attend} to corresponding epipolar positions in the other volume.
This is done by penalizing attention values that have high values outside the epipolar lines, and encouraging the attention along epipolar lines to be high. This is achieved by using a Binary Cross Entropy (BCE) loss on the \textit{raw} cross-attention maps $\{A^{12},A^{21}\}$:
\vspace{-6pt}
\begin{align}
L^{12}(i,j)&=\text{BCE}(\sigma(A^{12}(i,j)),\mathbbm{1}(i,j)) \nonumber \\
L^{21}(i,j)&=\text{BCE}(\sigma(A^{21}(i,j)),\mathbbm{1}(i,j)) \nonumber \\
L_{EPI} &= \sum_{i=1}^{s^2}\sum_{j=1}^{s^2} L^{12}(i,j)+L^{21}(i,j) \label{eqn:epiloss}
\end{align}
where $\sigma$ is a sigmoid function, and $\mathbbm{1}(i,j)$ is a special indicator function that is $1$ when location $j$ in the other feature map lies on the epipolar line corresponding to location $i$ in the current map.
The training process is illustrated in Fig.~\ref{fig:mainfig}.
\vspace{-20pt}
\paragraph{Max-Epipolar Loss.} In the Epipolar Loss proposed above, every point on the corresponding epipolar line is encouraged to have high attention even if it is not the \textit{actual} matching point in 3D. We also propose a variant called \textit{Max-Epipolar Loss}, wherein we select only the point on the epipolar line with the maximum predicted cross-attention value and encourage the attention for that point to be high.
\begin{equation}
L_{MaxEPI} = L_{zero}+L_{max} \label{eqn:maxepiloss}
\end{equation}
where
\begin{align}
L_{max} &= \sum_{i} \text{BCE}\left(\max_{j\in e_i} \sigma(A(i,j)),1\right) \nonumber \\
L_{zero} &= \sum_{\forall i,j,\mathbbm{1}(i,j)=0} \text{BCE}(\sigma(A(i,j)),0) \nonumber
\end{align}
where $e_i$ is the set of locations in the other feature map that lie on the epipolar line corresponding to location $i$ in the current map. $L_{zero},L_{max}$ are applied to both $A^{12}$ and 
$A^{21}$.

Note that the epipolar loss is applied at training time, so epipolar geometry is required only during training. The epipolar geometry can be obtained from the relative pose between the images, or from the images directly. However, as will be shown in Sec.~\ref{sec:discuss}, once trained with our proposed $L_{EPI}$, the attention map extracted from the trained RRT model for a previously \textit{unseen} pair of images shows patterns corresponding to the actual epipolar lines (\textit{without} any input epipolar geometry information). This demonstrates that the model's predictions are epipolar-geometry-aware, and at test time this leads to improved reranking performance as erroneous point matches can be avoided.

\subsection{Epipolar Positional Encoding}
\label{sec:epiposenc}
In contrast to Epipolar Loss where we implicitly induce awareness of epipolar line correspondence into the model, epipolar constraints can be encoded \emph{explicitly} by annotating each pixel 
with an encoding that uniquely identifies its epipolar plane.
The family of epipolar planes ``rotates" about the line joining the two camera centers, hence it can be parameterized by a scalar angle of rotation.
Inspired by \cite{Yifan_2022_CVPR}, we introduce a baseline wherein we encode the epipolar plane angle for each token and add 
the encoding to the input tokens of the transformer.
A random epipolar plane corresponding to a randomly chosen pixel location is used as reference to calculate the plane angle. We encode the angle with the frequency positional encoding \cite{vaswani2017attention,NeRF}. 

The drawback of the explicit method is that it requires the epipolar geometry (or relative pose) information during inference. In the scenario when this information is not available, we have to rely on other ways to obtain the epipolar geometry or relative pose which may not be entirely accurate and leads to loss in performance, as will be shown in Sec.~\ref{sec:res}. In fact, determining whether epipolar geometry can be established between two views is essentially replacing the job of the reranking transformer in determining if two images contain the same object.

\section{CO3D-Retrieve benchmark}
\label{sec:co3d-retrieve}
We now describe how we repurpose the 
CO3Dv2~\cite{co3d} dataset to create a large-scale object instance retrieval benchmark with multiple views of real objects.
CO3Dv2 is a dataset of multi-view images of common object categories, consisting of 36,506
videos of object instances (one video per object instance), taken from distant viewpoints spanning all 360 degrees, and covering 51 common object categories. 
The dataset also contains the ground-truth camera poses for the video frames and foreground segmentation masks for the object in each image.

\begin{figure*}
    \centering
    \includegraphics[width=0.8\textwidth,trim={2.0cm 0 0 0},clip]{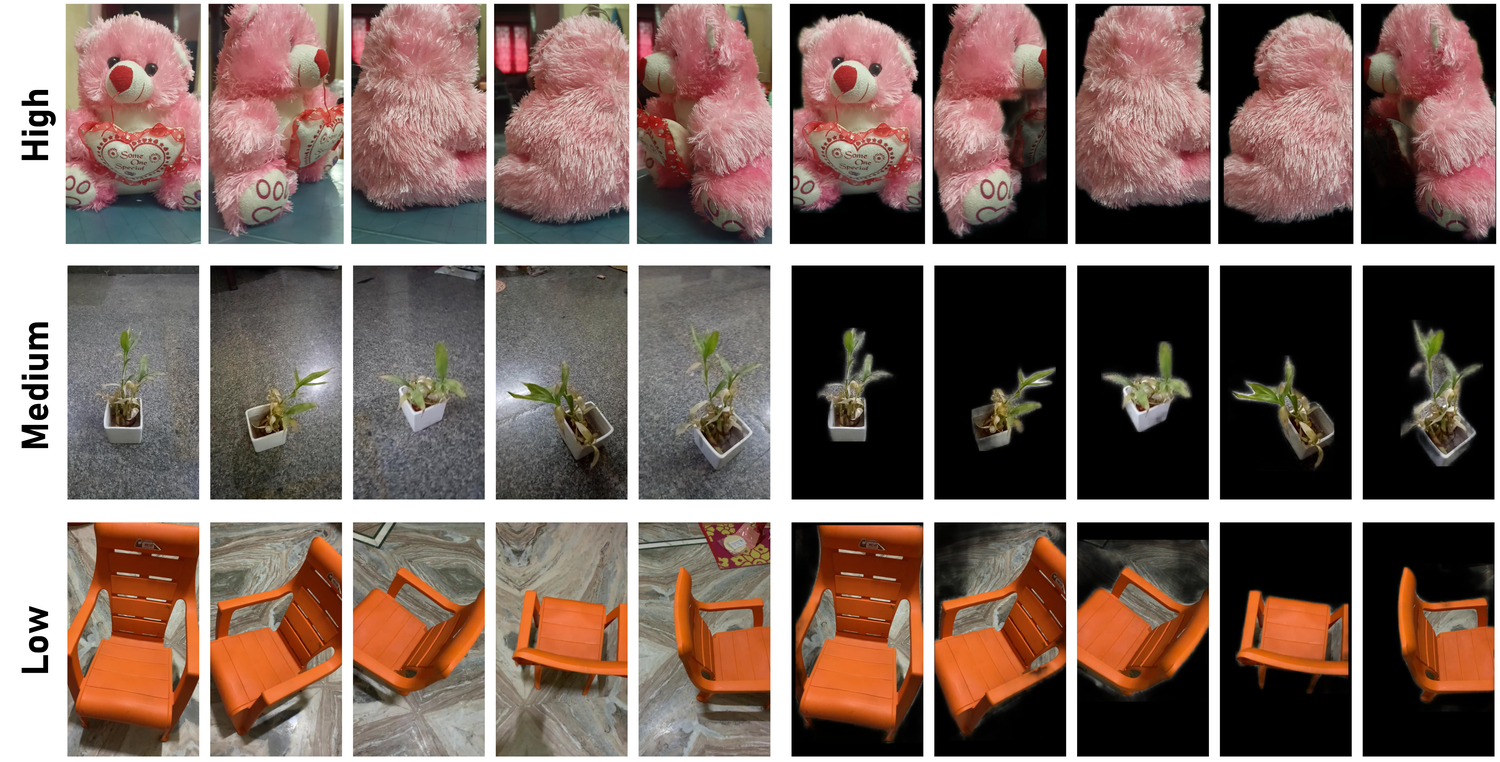}  %
    \caption{Example images for three object instances from the CO3D-Retrieve benchmark. The left half shows the full image, and the right half shows the masked counterparts  obtained using the object masks in CO3D \cite{co3d}. The number of pixels in common between views of the same object decreases from the top row to the bottom (computed using the 3D point-clouds, also available from CO3D).}
    \label{fig:co3d-retrieve}
    \vspace*{-10pt}
\end{figure*}

For the {\em CO3D-Retrieve} dataset, we extract 5 frames per video so that each frame is separated from the next by \textit{approximately} $72^{\circ}$ of rotation around the object.
In total, CO3D-Retrieve contains 181,857 images of 36,506 object instances. We split the dataset into two halves for training and testing: the training dataset contains 91,106 images of 18,241 object instances, and the testing datasets contains 90,751 images from 18,265 object instances. The set of object instances seen during training and testing are disjoint and so have zero overlap with each other. For benchmarking object retrieval on CO3D-Retrieve, we evaluate with each image as the query, the other images from the same object as the query are treated as \textit{positives}, and all the images not corresponding to the query object are treated as negatives. Fig.\ \ref{fig:co3d-retrieve} shows example object images from the benchmark. 

\section{Experiments}
\label{sec:exp}

In this section, we evaluate the  epipolar-geometry aware Reranking Transformer on two datasets: our CO3D-Retrieve benchmark, and the Stanford Online Products (SOP) benchmark~\cite{oh2016deep}. SOP is a popular benchmark for object retrieval containing 120,053 images of 22,634 object instances from 12 object categories. We use the standard train-test split used by all the baselines we compare with, where 59,551 images are used for training and 60,502 for testing. 
In Sec.~\ref{sec:discuss}, we provide a discussion on the merits of the implicit approach to incorporating epipolar constraints and explore its properties.

\subsection{Baselines and metrics}
\label{sec:baselines}
\noindent \textbf{Pretrained descriptors.}\ Deep networks pretrained on large scale image datasets learn powerful image representations that can be used for retrieval. Evaluating such pretrained models without fine-tuning gives us a lower bound on the performance that a model trained on our dataset should achieve. We compare with VGG16 \cite{Simonyan15} and ResNet50 (R50) \cite{ResNet} models pretrained on ImageNet~\cite{deng2009imagenet}, i.e.\ trained for classification, not retrieval.
We also compare to a NetVLAD~\cite{arandjelovic2016netvlad} model (i.e.\ VGG16 backbone + NetVLAD pooling) pretrained for retrieval on Pittsburgh250k \cite{torii2013visual}. \\

\noindent \textbf{Reranking Transformers (RRT).}
Reranking Tranformers (RRT) \cite{rrt} is a state-of-the-art method that our works builds on. We compare with different versions of the RRT method:
\begin{enumerate}
    \item \textit{R50 (trained)}: this baseline performs global retrieval (no reranking) and does not use RRT, but works as a foundation for subsequent baselines. The model is trained  using a batch-wise contrastive loss on CO3D-Retrieve or SOP \cite{oh2016deep}, for the respective experiments.
    \item \textit{R50 (frozen) + RRT}: we start from a trained R50 (i.e.\ baseline (1)), freeze its weights and train a RRT on top of it for reranking.
    \item \textit{R50 (finetune) + RRT}: we start from a trained R50 (i.e.\ baseline (1)) and we finetune the R50 backbone along with the RRT.
\end{enumerate}

\noindent \textbf{RRT w/ Epipolar Positional Encoding.} The R50 backbone along with RRT is trained with their respective ``retrieval loss" functions, and the epipolar geometry is provided as input in the form of an Epipolar Positional Encoding (Sec.\ \ref{sec:epiposenc}). We will discuss the results of this baseline in a separate Sec.\ \ref{sec:implexpl}. \\

\noindent \textbf{Evaluation metrics.}
Given a query image and a retrieved image, they match if they contain the same object instance. We report two metrics to evaluate retrieval performance. First, \textbf{R@K} -- 
for a given query, if a match is within the top $K$ retrieved images, then the query is said to be retrieved. $R@K$ is the fraction of correctly retrieved queries for a given $K$. We report results for $K=1$, $10$ and $50$.
Second, we report \textbf{Mean Average Precision (mAP)} --  the mean of the Average Precision \cite{10.5555/1394399} over all queries.

\subsection{Implementation details}
\label{sec:impl}

\noindent \textbf{Extracting the epipolar geometry.} For experiments with the CO3D-Retrieve benchmark, the available ground-truth pose information for each image is used to compute the epipolar geometry for a matching pair of images. During training, we use the ``RandomCrop" image augmentation which shifts the principal point location. Hence, we adjust the Fundamental Matrix computation appropriately to obtain the correct epipolar lines in the cropped images. When pose information is not available as ground-truth, such as in the SOP \cite{oh2016deep} dataset, we use an off-the-shelf method to compute the epipolar geometry. Specifically, we use a local image feature matching method, LoFTR \cite{sun2021loftr}, to extract high-quality semi-dense matches between the image pair. Then, we use a robust estimation method, MAGSAC++ \cite{barath2020magsac++} to extract the Fundamental Matrix. The epipolar geometry extracted with this method is not entirely accurate (especially for image pairs with extreme relative pose), but it provides us with a sufficient pseudo ground-truth epipolar geometry to train our models with Epipolar Loss. If the number of matches found $\leq 20$ or number of inliers detected $\leq 0.2\times$ number of matches, we consider the extracted epipolar geometry unreliable and do not apply Epipolar Loss for that image pair during training. Computing epipolar geometry with this method takes $\approx\!0.06$ seconds per image pair on a 8-core CPU and NVIDIA P40 GPU. 

\noindent \textbf{Training details.} We use a ResNet50~\cite{ResNet} for global retrieval, which is trained with a batchwise contrastive loss (batch size of $800$).  For an image pair\{$\mathcal{I}$, $\mathcal{\bar{I}}$\} in the batch,  a Binary Cross-Entropy loss is used to train the reranking model enforcing its output to be $1$ if $\mathcal{I}$ and $\mathcal{\bar{I}}$ contain the same object and $0$ otherwise. 

In our proposed implicit method, the global retrieval model and the reranking model are trained with their respective retrieval-losses plus the Epipolar Loss. If $\mathcal{I}$ and $\mathcal{\bar{I}}$ represent the same object, then the epipolar geometry between the image pair (which is extracted as explained above) is used to compute the Epipolar Loss for training. If the image pair is not a match, then a valid epipolar geometry does not exist 
and we simply do not apply the Epipolar Loss for that image pair. 

In the explicit method, we have to include the geometry in the input as Epipolar Positional Encodings (EPE), even when the input pair \{$\mathcal{I}$, $\mathcal{\bar{I}}$\} is not a match. To handle the case when \{$\mathcal{I}$, $\mathcal{\bar{I}}$\} is not a match during training and testing, we use a \textit{random} rank-2 matrix as the Fundamental Matrix to compute the EPEs. When \{$\mathcal{I}$, $\mathcal{\bar{I}}$\} is indeed a match, (a) during training, we use the ground-truth or the pseudo ground-truth (whichever is available) to compute the EPEs, (b) during testing, we do not rely on the ground-truth geometry information and always use the LoFTR/MAGSAC++ method (described above) to compute the EPEs.

\begin{table*}[!t]
\centering
\caption{Evaluation on CO3D-Retrieve benchmark. Description of all compared methods in Sec.\ \ref{sec:baselines}. Baselines shown above dashed line are pretrained models and below are trained on CO3D-Retrieve. \textbf{EPE}=Epipolar Positional Encoding}
\label{tab:co3d-results}
\begin{tabular}{lcccccccc} 
\toprule 
\multirow{ 2}{*}{Method} & \multicolumn{4}{c}{Full images} & \multicolumn{4}{c}{With masked backgrounds} \\
\cmidrule(l{4pt}r{4pt}){2-5} \cmidrule(l{4pt}r{4pt}){6-9}
& $R@1$ & $R@10$ & $R@50$  & mAP & $R@1$ & $R@10$ & $R@50$  & mAP \\
\midrule
\hspace{-5pt}\textit{Pretrained Models} & & & & & & & & \\
VGG16 \cite{Simonyan15}  & 66.21 & 85.18 & 91.66 & 22.51 & 63.56 & 81.11 & 89.24 & 16.73 \\
R50 \cite{ResNet}  & 66.48 & 85.34 & 91.74 & 22.79    &  63.81 & 80.30 & 89.37 & 16.79 \\
NetVLAD \cite{arandjelovic2016netvlad} & 67.01 & 85.17 & 91.72 & 22.63 & 63.19 & 80.84 & 89.27 & 16.61 \\[0.5ex]
\hdashline
\noalign{\vskip 0.5ex}
R50 (trained) &  86.06 & 95.62 & 97.65 & 45.34 &  78.82 & 91.30 & 94.72 & 24.85 \\
RRT \cite{rrt}
+ R50 (frozen)  &  88.07 & 96.29 & 97.75 & 47.60 & 82.45 & 91.89 & 94.85 & 26.16 \\
RRT
+ R50 (finetune) (\textbf{SOTA}) &  89.20 & 96.85 & 97.89 & 48.81 &  83.28 & 92.13 & 95.05 & 27.33 \\
RRT + R50 w/ EPE & 88.53 & 96.41 & 97.83 & 47.99  & 82.79 & 91.96 & 94.99 & 26.58 \\
RRT + R50 w/ $L_{EPI}$ (\textbf{Ours}) &  90.57 & 97.33 & 98.10 & 49.52 &  85.07 & 92.42 & 95.11 & 28.07 \\
RRT + R50 w/ $L_{MaxEPI}$ (\textbf{Ours}) & \textbf{90.69} & \textbf{97.38} & \textbf{98.10}  & \textbf{49.60} & \textbf{85.17} & \textbf{92.46} & \textbf{95.14} & \textbf{28.21} \\
\bottomrule 
\end{tabular}
\end{table*}

The hyperparameters we use for our experiments with SOP \cite{oh2016deep} are the same as \cite{rrt}, except that we use 40 epochs (instead of 100) when training with the Epipolar Loss with a constant learning rate of $10^{-4}$. Hyperparameters used with CO3D-Retrieve are provided in \cref{sec:app-impl}.
Our method is entirely implemented in PyTorch \cite{paszke2019pytorch}.

\subsection{Results on CO3D-Retrieve}
\label{sec:res}
We evaluate on CO3D-Retrieve in two settings: with and without masking the background in the object images. This is because the background also provides useful visual cues for image matching and it is essential to see how the methods perform without any such extra information.  Figure~\ref{fig:co3d-retrieve} shows examples with and without masking the background.

Table \ref{tab:co3d-results} shows the detailed results.
We observe that pretrained models (VGG16 and R50 on ImageNet, NetVLAD on Pittsburgh250k) achieve a reasonable performance without any finetuning. However, their performance is not competitive compared with baselines specialized for CO3D-Retrieve. It's interesting to note that a simple ResNet50-based global retrieval baseline trained with batch-wise contrastive loss (i.e.\ the ``R50 (trained)" baseline) already achieves a high $R@1$ of $86.06\%$ on the unmasked images.  Our method, which uses the Epipolar Loss to induce multi-view geometric understanding in to the Reranking Transformer model, outperforms the state-of-the-art approach (RRT + R50 (finetune)) in both masked and unmasked settings. The margin with which the Epipolar Loss baseline outperforms ``RRT + R50 (finetune)" is greater in the case of images with masked background, as this is a harder task. 
It can be seen that the Max-Epipolar variant of the Epipolar loss consistently gives a slight improvement.

\begin{table} %
    \centering
      \setlength{\tabcolsep}{2pt}
      \caption{Evaluation on Stanford Online Products \cite{oh2016deep}. Baselines shown above the dashed line are pretrained models and below the dashed line are trained on SOP. More details in Sec.~\ref{sec:baselines}. 
      Key: \scriptsize{*=results obtained using checkpoints from~\cite{rrt}};  \scriptsize{**=results reported in~\cite{rrt}}}
      \label{tab:sop-results}
        \begin{tabular}{>{\small}lcccc} 
\toprule 
Method & $R@1$ & $R@10$ & $R@50$  & mAP\\
\midrule
\hspace{-2pt}\textit{Pretrained Models} & & & & \\
VGG16 \cite{Simonyan15} &  55.75 & 70.86 & 79.65 & 11.93 \\
R50 \cite{ResNet} & 55.89 & 71.32 & 79.69 & 12.09 \\
NetVLAD \cite{arandjelovic2016netvlad} & 54.16 & 70.85 & 79.62 & 11.90 \\[0.5ex]
\hdashline
\noalign{\vskip 0.5ex}
R50 (trained)* &  80.74 & 91.87 & 95.54 & 32.90 \\
RRT \cite{rrt}
+ R50 (frozen)* &  81.80 & 92.35 & 95.78 & 34.91 \\
RRT
+ R50 (finetune)* (\textbf{\small SOTA}) &  84.46 & 93.21 & \textbf{96.04} & 37.14 \\
RRT 
+ R50 w/ EPE &  82.57 & 92.69 & 95.89 & 35.38 \\
RRT
+ R50 w/ $L_{EPI}$ (\textbf{\small Ours}) & \textbf{84.74} & \textbf{93.29} & \textbf{96.04} & \textbf{37.25} \\
RRT
+ R50 w/ $L_{MaxEPI}$ (\textbf{\small Ours}) & 84.53 & 93.27 & \textbf{96.04} & 37.19 \\
\midrule
\hspace{-2pt}\textit{Other Metric Learning methods**} & & & \\
Margin-based \cite{roth2020revisiting} & 76.1 & 88.4 & - & - \\
FastAP \cite{cakir2019deep} & 73.8 & 88.0 & - & -\\
XBM \cite{wang2020cross} & 80.6 & 91.6 & - & - \\
Cross-Entropy based \cite{boudiaf2020metric} & 81.1 & 91.7 & - & -\\
\bottomrule
\end{tabular}
\end{table}

\begin{table} %
    \centering
        \setlength{\tabcolsep}{2pt}
        \caption{Zero-shot evaluation on Stanford Online Products \cite{oh2016deep} with models trained on CO3D-Retrieve.}
        \label{tab:zeroshot}
        \begin{tabular}{>{\small}lcccc} 
\toprule 
Method & $R@1$ & $R@10$ & $R@50$  & mAP\\
\midrule
RRT + R50 (frozen) & 75.53 & 89.43 & 95.01 & 29.27 \\
RRT + R50 (finetune) & 76.32 & 90.16 & 95.21 & 30.19 \\
RRT + R50 w/ $L_{EPI}$ (\textbf{Ours}) & \textbf{76.78} & \textbf{90.27} & \textbf{95.29} & \textbf{30.25}\\
\bottomrule
\end{tabular}
\end{table}

\subsection{Results on Stanford Online Products}
The Stanford Online Products (SOP) dataset \cite{oh2016deep} does not contain ground-truth pose information for the object images. As detailed in Sec.\ \ref{sec:impl}, we obtain the pseudo ground-truth geometry information using LoFTR \cite{sun2021loftr} for matching and MAGSAC++ \cite{barath2020magsac++} for robust estimation. We find that, even though these pseudo ground-truth poses are not entirely accurate, they are still useful for training with the Epipolar Loss.
Table \ref{tab:sop-results} shows a comprehensive comparison of our proposed methods with all the baselines.
We also include deep metric learning methods \cite{roth2020revisiting, cakir2019deep, wang2020cross, boudiaf2020metric} with reported numbers taken from \cite{rrt} into our comparisons. Our proposed implicit method outperforms all the baselines including the state-of-the-art Reranking Transformers \cite{rrt}.

We also test zero-shot retrieval, by evaluating on SOP models that were trained on CO3D-Retrieve.
The results are shown in Table~\ref{tab:zeroshot}, where we can observe that the differences between all methods are reduced, but our Epipolar Loss still confers a performance advantage.

\begin{figure}[!t]
    \centering
    \includegraphics[width=\columnwidth]{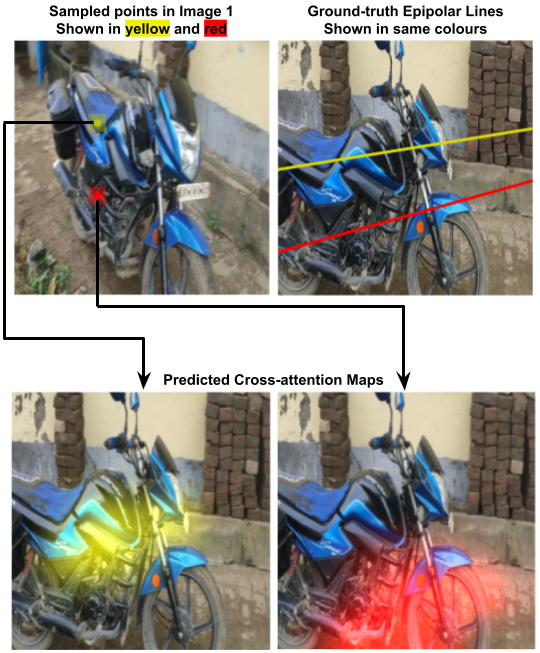}
    \caption{Cross-attention maps, extracted from the RRT model trained with Max-Epipolar Loss, overlaid on the image. Note: the attention maps are bilinearly upsampled to the size of the image.}
    \label{fig:maxepi-overlay}
\end{figure}

\begin{figure}[!t]
    \centering
    \includegraphics[width=0.95\linewidth]{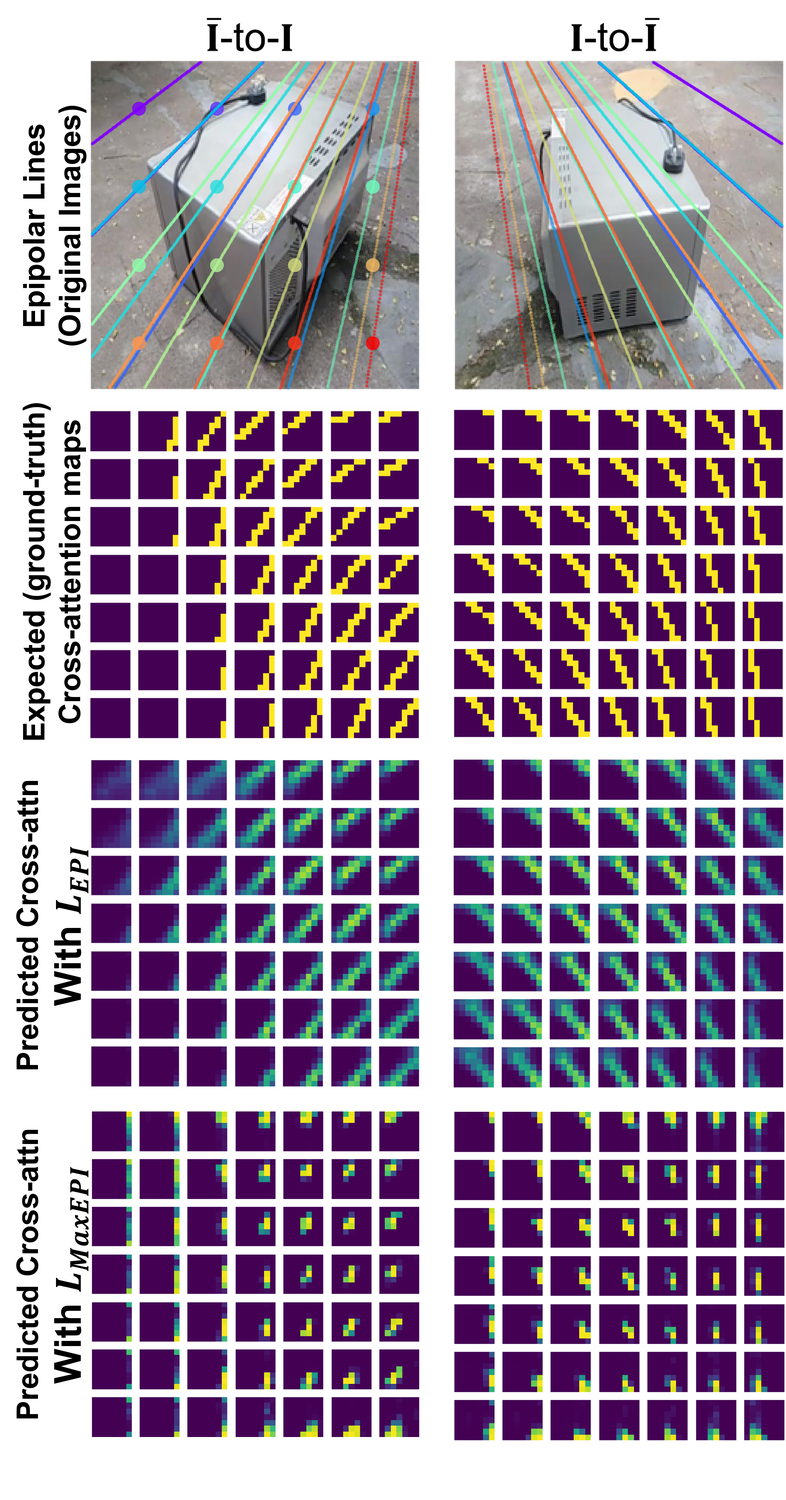}
    \caption{Visualization for a test image pair (i.e.\ never seen in training).
    \textbf{Top row}: Points shown in image $\mathcal{I}$ have correspondences on the epipolar lines of that color in image $\mathcal{\bar{I}}$. \textbf{Second row}: Expected $7\!\!\times\!\!7\!\!\times\!\!7\!\!\times\!\!7$ cross-attention maps shown as a $7\!\!\times\!\!7$ grid with a $7\!\!\times\!\!7$ patch at each grid location computed from the ground truth epipolar geometry. In the $\mathcal{I}$-to-$\mathcal{\bar{I}}$ grid (right column), a patch at grid location ($i,j$) shows the epipolar line in $\mathcal{\bar{I}}$ corresponding to the pixel ($i,j$) in the $7\!\!\times\!\!7$ feature space of $\mathcal{I}$. \textbf{Third row}: Predicted cross-attention maps from transformer trained with $L_{EPI}$. Notice how closely they match ground-truth maps, even though these are test images and do not have access to the ground truth epipolar geometry.
    \textbf{Bottom row}: Predicted cross-attention maps from transformer trained with Max-Epipolar Loss, $L_{MaxEPI}$. These are sparser and have peaks that lie close to actual epipolar lines.
    }
    \label{fig:vizattn_full}
\end{figure}

\subsection{Implicit vs Explicit methods}
\label{sec:implexpl}
The transformer model trained with Epipolar Loss does not require pose or epipolar geometry information at test time. The explicit method, however, requires the fundamental matrix at the input (during both training and testing) to generate the Epipolar Positional Encodings (EPE). 
Tables~\ref{tab:co3d-results} and~\ref{tab:sop-results} show that using EPE with Reranking Transformer adversely affects the performance, compared to not using the encodings.
Although reasons for this decrease are unclear, one possibility is that the encodings leak information about whether two images match or not, because when they do not match, the input epipolar encodings are arbitrary. The network may learn to rely on this signal instead of image matching, in a case of ``shortcut learning''~\cite{geirhos2020shortcut}.
This issue does not affect the implicit method as geometry information isn ot required at test time.\ When training the RRT, the Epipolar Loss is used with \textit{only} those image pairs that contain matching images. Hence, during inference, the Transformer uses implicit geometry information only when it is \textit{valid} (i.e.\ inherently for matching image pairs).

\subsection{What does the implicit model learn?}
\label{sec:discuss}

After training with the Epipolar Loss ($L_{EPI}$) or the Max-Epipolar Loss ($L_{MaxEPI}$), we investigate if the attention maps of the learned model show some signs of geometric-awareness. To do this, we pick two matching images \{$\mathcal{I},\mathcal{\bar{I}}$\} from the \textit{test} set, i.e.\ these images were not seen during training, and extract the cross-attention maps from the last layer of the transformer. Since we use a $7\times7\times128$ feature volume for each image (reshaped to $49\times128$ for the transformer), the cross-attention maps (from $\mathcal{I}$ to $\mathcal{\bar{I}}$, and $\mathcal{\bar{I}}$ to $\mathcal{I}$) are of size $49\times49$ which are then reshaped back to $7\times7\times7\times7$. These $7\times7\times7\times7$ cross-attention map values indicate the attention between each \textit{feature-pixel} of the first and second feature volume.

Fig. \ref{fig:vizattn_full} shows predicted cross-attention maps alongside expected ground-truth maps. The latter are computed using ground-truth pose information. We observe that the attention maps obtained with $L_{EPI}$ closely follow ground truth epipolar lines, despite this instance and associated geometry not being seen during training. Note that the attention maps obtained with $L_{MaxEPI}$ are much sparser with peaks that lie on actual epipolar lines. \\

\noindent \textbf{Do cross-attention maps learned with Max-Epipolar Loss correspond to true matching points?} We overlay the cross-attention maps, extracted from the RRT \cite{rrt} model trained with Max-Epipolar Loss ($L_{MaxEPI}$), on the original images to see if the location of highest attention coincides with the position of the actual matching point. \cref{fig:maxepi-overlay} shows one such visual illustration. We can see that the peaks of attention maps \textit{very loosely} coincide with the actual matching points on the corresponding ground-truth epipolar lines.

\begin{figure}
    \centering
    \includegraphics[width=0.8\linewidth]{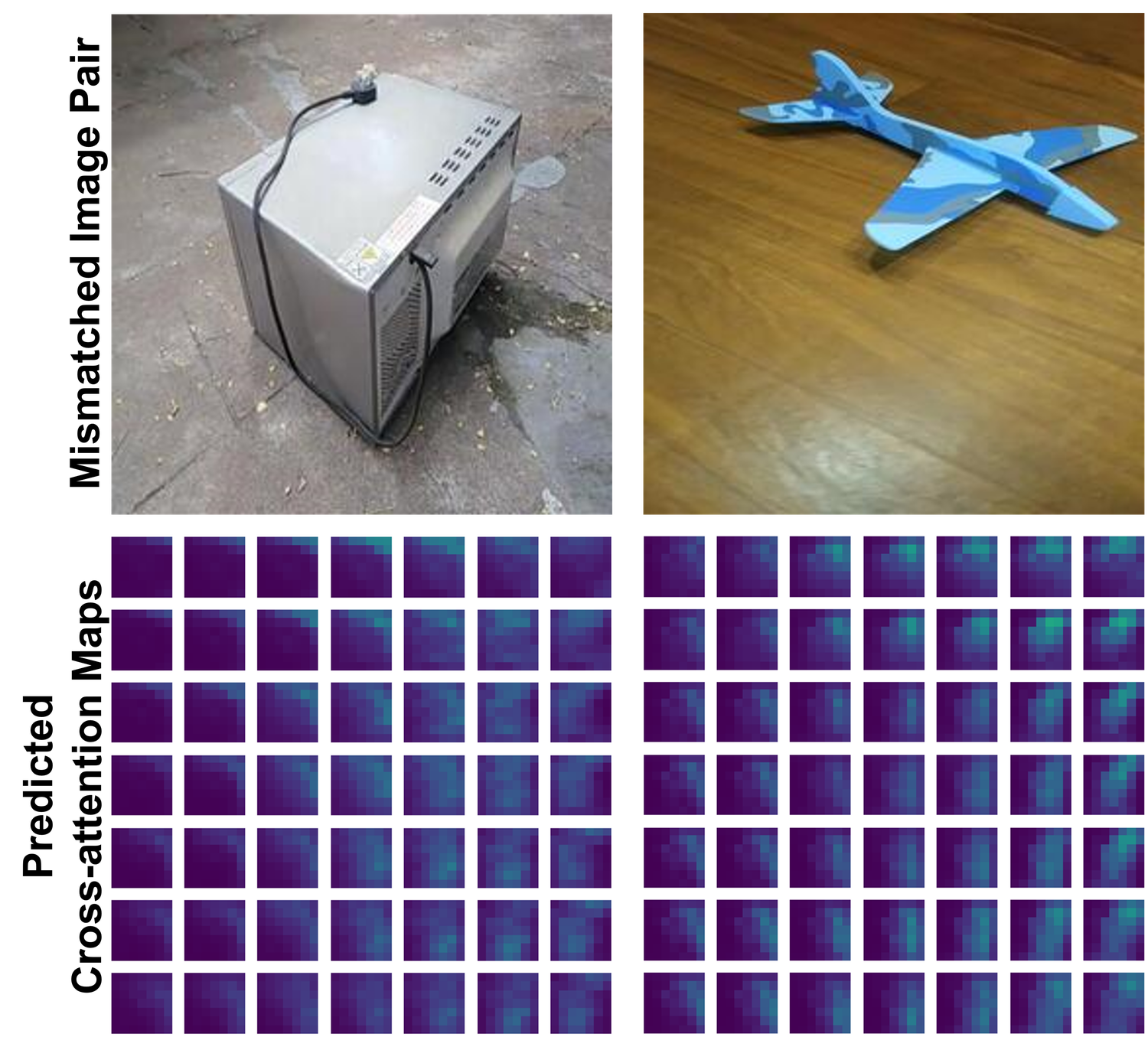}
    \caption{Predicted attention maps for a non-matching test image pair. A valid epipolar geometry does not exist for this pair, hence the model predicts \textit{diffuse} attention maps.}
    \label{fig:vizattn_mis}
\end{figure}

Additionally, we visualize the cross-attention maps for a pair of \textit{mismatched} images, as shown in \cref{fig:vizattn_mis}. These cross-attention maps are extracted  from a RRT model trained with Epipolar Loss ($L_{EPI}$).

\vspace{-2pt}
\section{Conclusion}
In this work, we aimed to teach multi-view geometry to Transformer networks, and proposed a method to do so implicitly via epipolar guides.
The advantages of this implicit approach over explicitly passing in geometric information to a network are two-fold:
(i) ground-truth epipolar geometry (relative pose) between views is only needed at training time, not at inference;
(ii) implicit losses are readily applied to existing architectures, so there is no need to design specialized architectures.
We demonstrated improved performance over the state-of-the-art in object retrieval, by reranking with our method.
More generally, this approach of implicitly incorporating knowledge into Transformers by a suitable loss can be employed in other scenarios.
Examples include learning other geometric relations, such as a trifocal relationship over three views, as well as physical laws such as Newton's laws of motion.

\vspace{-6pt}
{\paragraph{Acknowledgements.}
We are grateful for funding from EPSRC AIMS CDT EP/S024050/1, AWS, the Royal Academy of Engineering (RF\textbackslash 201819\textbackslash 18\textbackslash 163), EPSRC Programme Grant VisualAI EP/T028572/1, and a Royal Society Research
Professorship RP$\backslash$R1$\backslash$191132. We thank the authors of~\cite{rrt,sun2021loftr,barath2020magsac++} for open-sourcing their code. We also thank an anonymous reviewer for useful suggestions on the Max-Epipolar Loss.}

{\small
\bibliographystyle{ieee_fullname}
\bibliography{bib/longstrings,bib/vgg_local,egbib}
}

\appendix




\section{Qualitative examples}
\begin{figure*}
    \centering
    \resizebox{\textwidth}{!}{%
    \includegraphics[height=6cm]{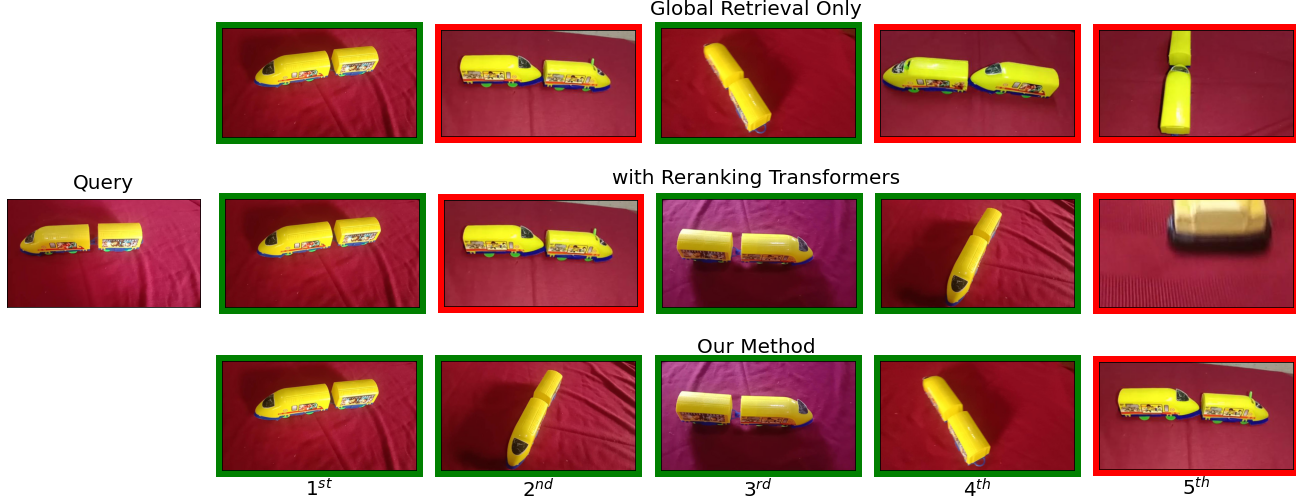}%
    \quad
    \includegraphics[height=6cm]{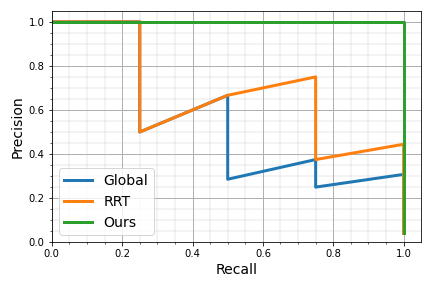}%
    }
    \caption{CO3D-Retrieve: Example 1.}
    \label{fig:co3d-ex1}
\end{figure*}

\begin{figure*}
    \centering
    \resizebox{\textwidth}{!}{%
    \includegraphics[height=6cm]{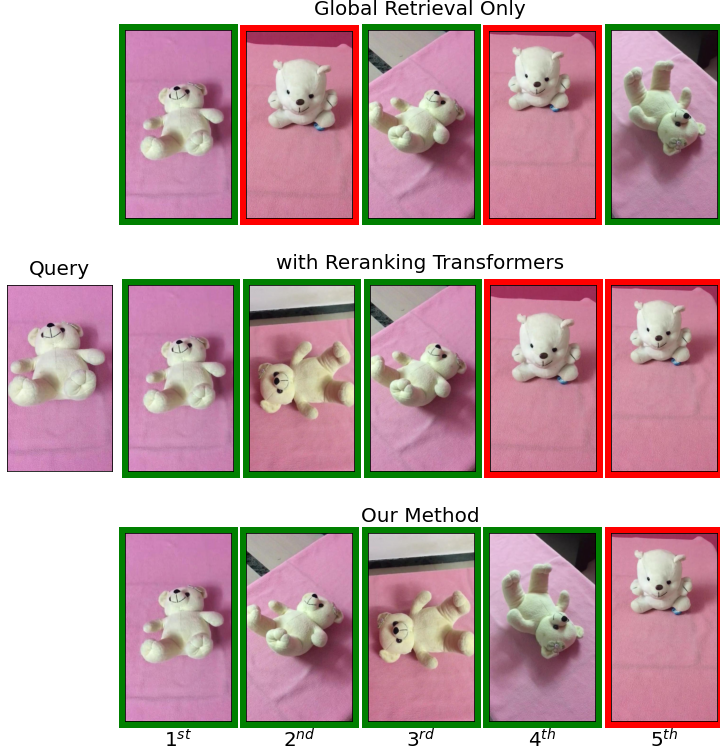}%
    \quad
    \includegraphics[height=4cm]{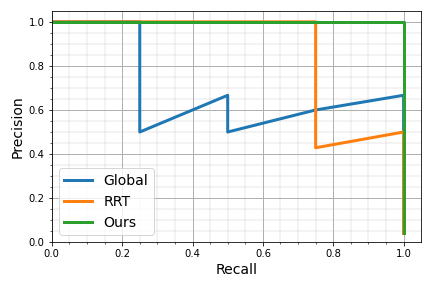}%
    }
    \caption{CO3D-Retrieve: Example 5.}
    \label{fig:co3d-ex5}
\end{figure*}

\begin{figure*}
    \centering
    \resizebox{\textwidth}{!}{%
    \includegraphics[height=6cm]{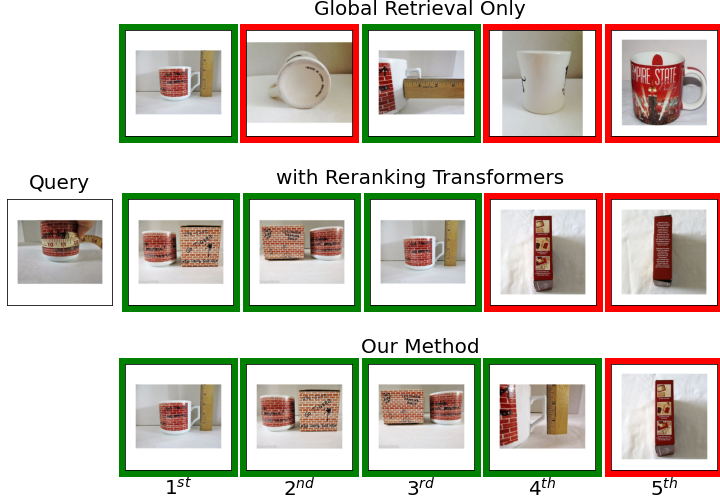}%
    \quad
    \includegraphics[height=5cm]{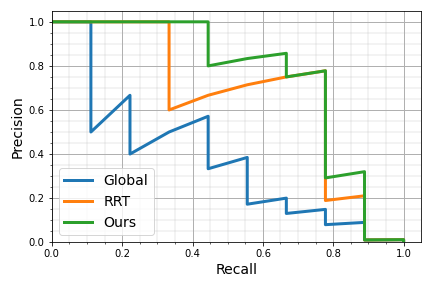}%
    }
    \caption{SOP \cite{oh2016deep} dataset. Example 1.}
    \label{fig:sop-ex1}
\end{figure*}

\begin{figure*}
    \centering
    \resizebox{\textwidth}{!}{%
    \includegraphics[height=6cm]{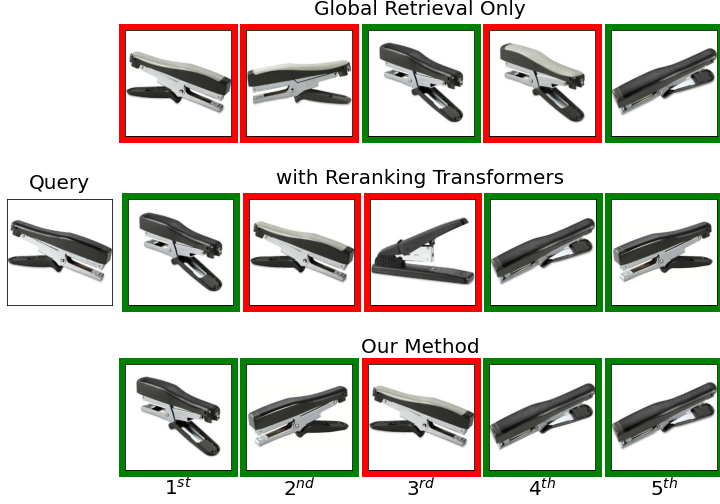}%
    \quad
    \includegraphics[height=5cm]{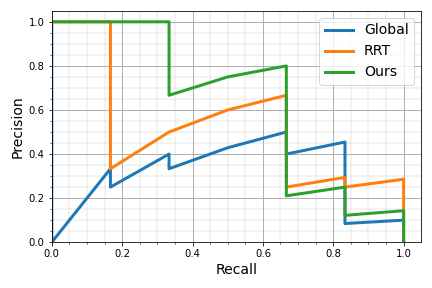}%
    }
    \caption{SOP \cite{oh2016deep} dataset. Example 6.}
    \label{fig:sop-ex6}
\end{figure*}

In Figures \ref{fig:co3d-ex1}-\ref{fig:sop-ex6}, we provide a few qualitative results on the CO3D-Retrieve benchmark (Figures \ref{fig:co3d-ex1}-\ref{fig:co3d-ex5}) and the Stanford Online Products \cite{oh2016deep} dataset (Figures \ref{fig:sop-ex1}-\ref{fig:sop-ex6}). We visually compare the top-$5$ retrievals obtained with the Global Retrieval (R50) model, Reranking Transformer \cite{rrt} model and a reranking model trained with our Epipolar Loss. We also accompany each example with its corresponding Precision-Recall curve, which provides a more detailed perspective on the retrieval performance.

In the CO3D-Retrieve benchmark, the \textit{maximum} number of reference images per query is $4$. So, for all the examples shown, the Precision-Recall curve for our method saturates at Precision $=1.0$ since the top-$4$ retrievals are correct.

\section{Visualization of Attention Maps}
\begin{figure}[!ht]
    \centering
    \begin{subfigure}{\columnwidth}
        \includegraphics[width=\columnwidth]{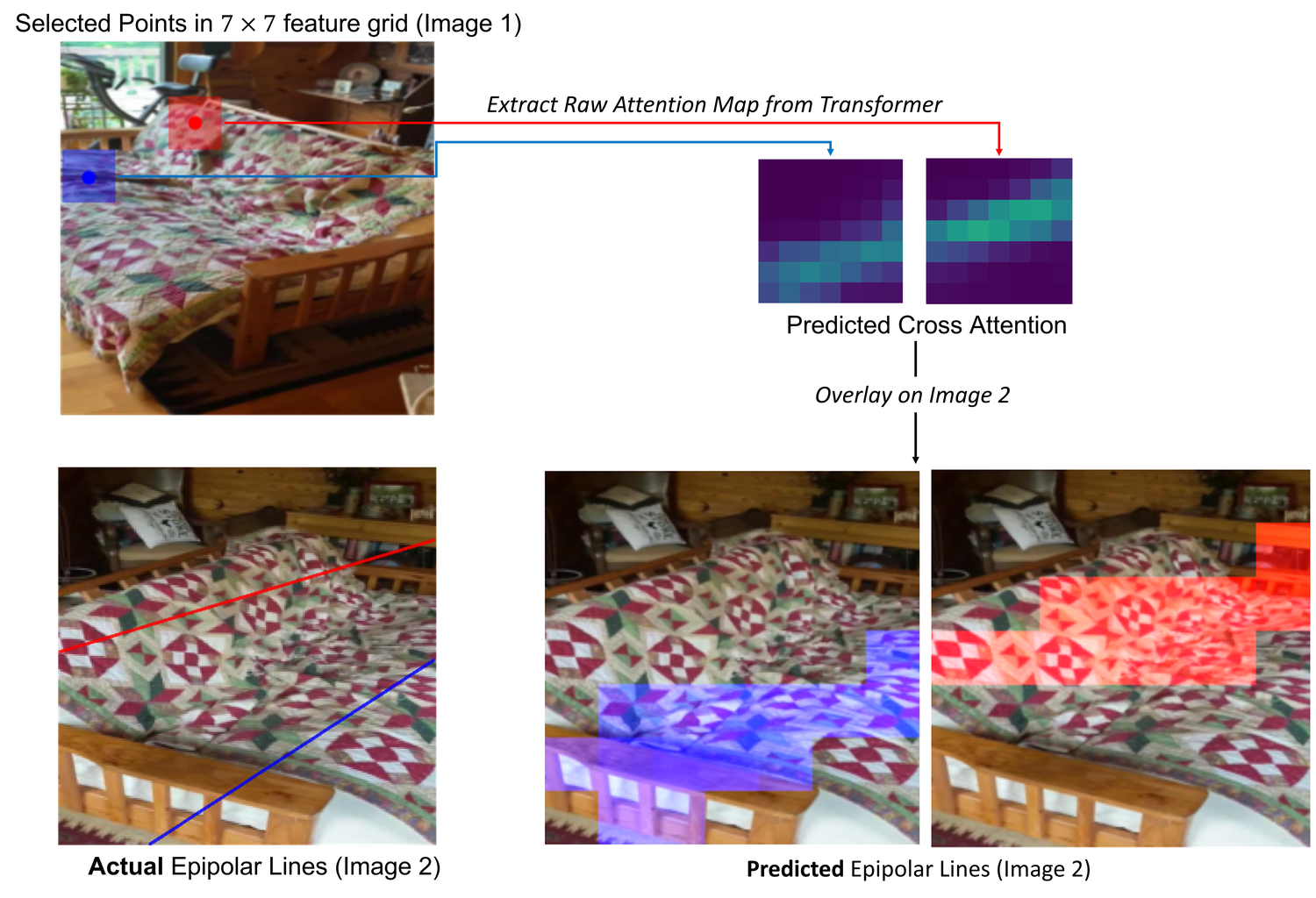}  
        \caption{}
        \vspace*{1cm}
    \end{subfigure}
    \begin{subfigure}{\columnwidth}
        \includegraphics[width=\columnwidth]{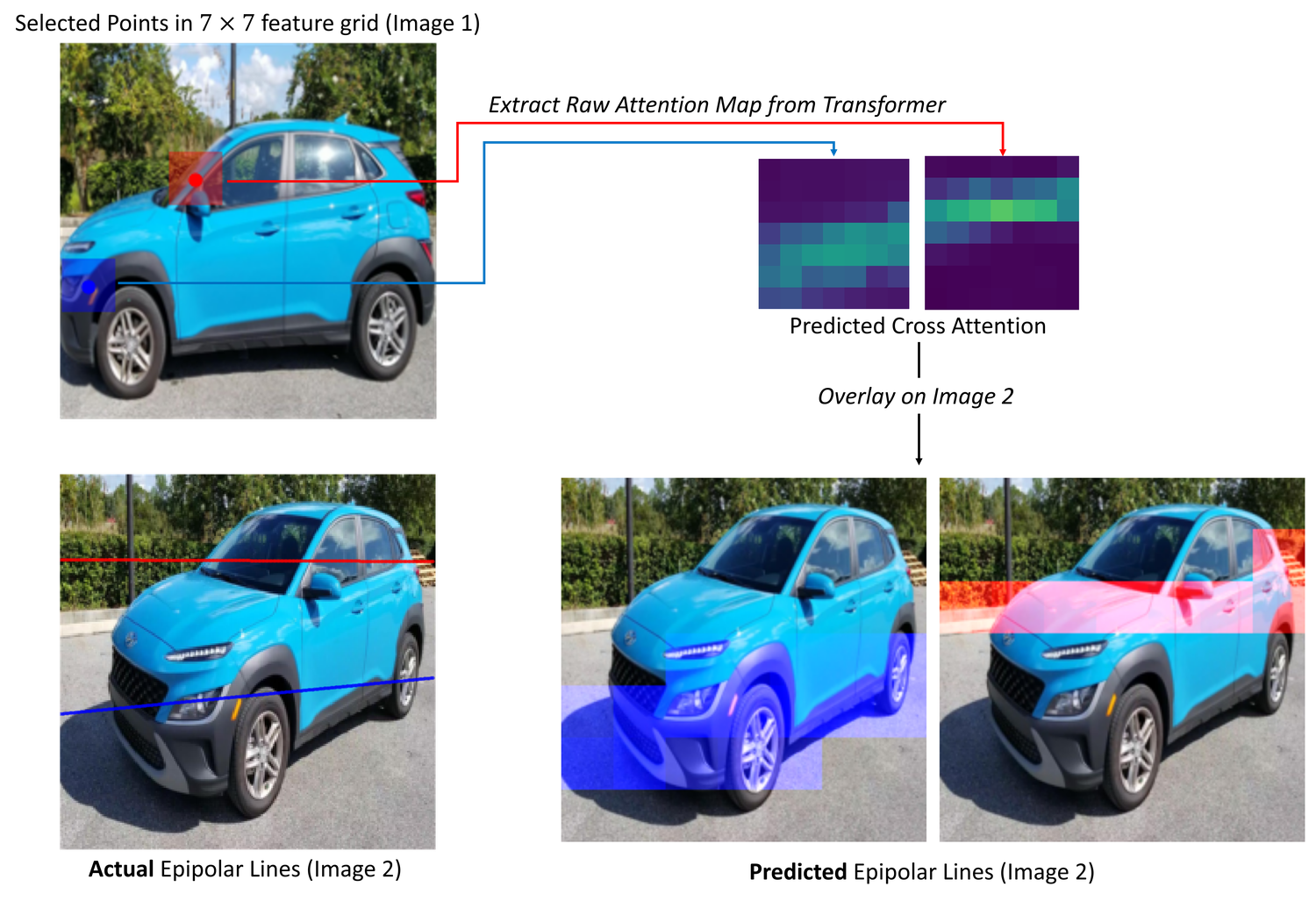}    
        \caption{}
    \end{subfigure}
    \caption{Examples showing how the cross-attention map predicted by our transformer model (trained with Epipolar Loss) contains information about the true epipolar geometry. \textcolor{red}{Red} and \textcolor{blue}{Blue} colours are used to show the two selected points and their corresponding actual vs predicted epipolar lines.}
    \label{fig:attn-viz-ex}
\end{figure}

For the sake of clarity, we describe, with an example, how the cross-attention map predicted by our Transformer model (trained with Epipolar Loss) contains information about the true epipolar geometry between the input image pair. Figure \ref{fig:attn-viz-ex} shows such an example, where we select two points in a $7\times7$ grid (because the feature map extracted by our backbone is spatially $7\times7$) of the first image and show the actual (ground-truth) as well as the predicted epipolar lines in the other image.

\section{Implementation details CO3D-Retrieve}
\label{sec:app-impl}
For experiments with the CO3D-Retrieve benchmark, the global-retrieval-only model (R50 (trained) as described in Sec.\ 5.1) is trained for 50 epochs with the Adam optimizer and a learning rate that starts at 0.0001 and decays exponentially by a factor of 10 every 20 epochs. The Reranking Transformer head is trained on top of this trained global model, by either freezing or finetuning the global model, with or without the Epipolar Loss. When training without the Epipolar Loss, the model is trained using a SGD optimizer with an initial learning rate of $5\times10^{-5}$ decayed exponentially by a factor of $10$ over $40$ epochs. When training with the Epipolar Loss, the above procedure is followed without Epipolar Loss for $20$ epochs and the model is trained for an additional $20$ epochs with Epipolar Loss with a learning rate of $10^{-6}$. As mentioned in \cref{sec:impl}, the hyperparameters we use for our experiments with SOP \cite{oh2016deep} are the same as \cite{rrt}, except that we use $40$ epochs (instead of $100$ in \cite{rrt}) when training with the Epipolar Loss with a constant learning rate of $10^{-4}$.

\section{mAP analysis with same category retrieval}
We empirically observe that a majority of high-ranked (i.e. top-5) false positives are images from the same category. We conduct an experiment where we compute the mAP while ranking only the images from the same category
as the query image and ignoring out-of-class images. As shown in \cref{table:same-cat-mAP}, we get a higher mAP when retrieval is only performed on the same category images. This means there are confusing images from \textit{outside} the categories, albeit a small fraction compared to intra-class.

Further per-category analysis with our method reveals that the \textit{top 5} categories with the highest proportions of intra-class false positives (in descending order) are \texttt{banana, suitcase, laptop, keyboard, umbrella} in CO3D-Retrieve dataset and \texttt{fan, cabinet, mug, coffee\_maker, kettle} in SOP \cite{oh2016deep}.

\begin{table}[t]
\centering
\begin{tabular}{lcc} 
\toprule 
Dataset Name & \makecell{Same Category \\Retrieval} & \makecell{Full Dataset \\Retrieval} \\
\midrule
CO3D-Retrieve & $52.03$ & $49.52$ \\
SOP \cite{oh2016deep} & $38.61$ & $37.25$ \\
\bottomrule 
\end{tabular}
\caption{Comparison of mAP computed while ranking only the images from the same category as the query image.}
\label{table:same-cat-mAP}
\end{table}

\section{Breakdown of $R@K$ based on fraction of overlapping pixels}
\begin{figure}
    \centering
    \includegraphics[width=\linewidth]{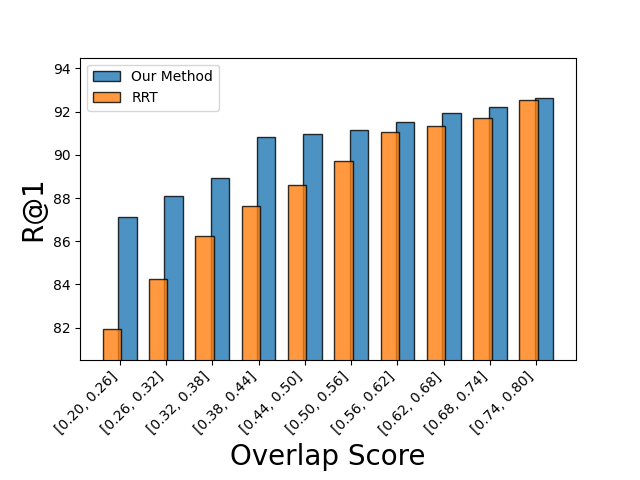}
    \caption{Breakdown of $R@1$ according to the Overlap Score of instances in the CO3D-Retrieve benchmark. The query set is divided into bins based on OS; these bins are shown on the x-axis.}
    \label{fig:breakdown}
\end{figure}

The proposed CO3D-Retrieve benchmark includes large variations in viewing angle among images of the same instance. We conduct an analysis to understand how our proposed method performs under a range of pose variations. To do this, we first compute an "Overlap Score (OS)"  for each instance using ground-truth point-clouds available in CO3D-Retrieve. Large pose-variations lead to a low OS. We divide the query-set into $10$ bins uniformly between OS=$0.2$ to $0.8$ and compute the $R@1$ for each bin. These limits of OS are chosen because there are very few ($<1\%$) instance with an OS beyond the $[0.2,0.8]$ range. \cref{fig:breakdown} shows $R@1$ for our proposed method and RRT \cite{rrt} with respect to the OS. We can see that the R@1 of our method drops by $5.5\%$ from highest to lowest OS, while RRT \cite{rrt} drops by $10.6\%$. In conclusion, our Epipolar Loss is useful in extreme viewpoint changes.

\section{High-resolution results}
\begin{figure}[!ht]
    \centering
    \begin{subfigure}{\columnwidth}
        \centering
        \includegraphics[width=0.9\columnwidth]{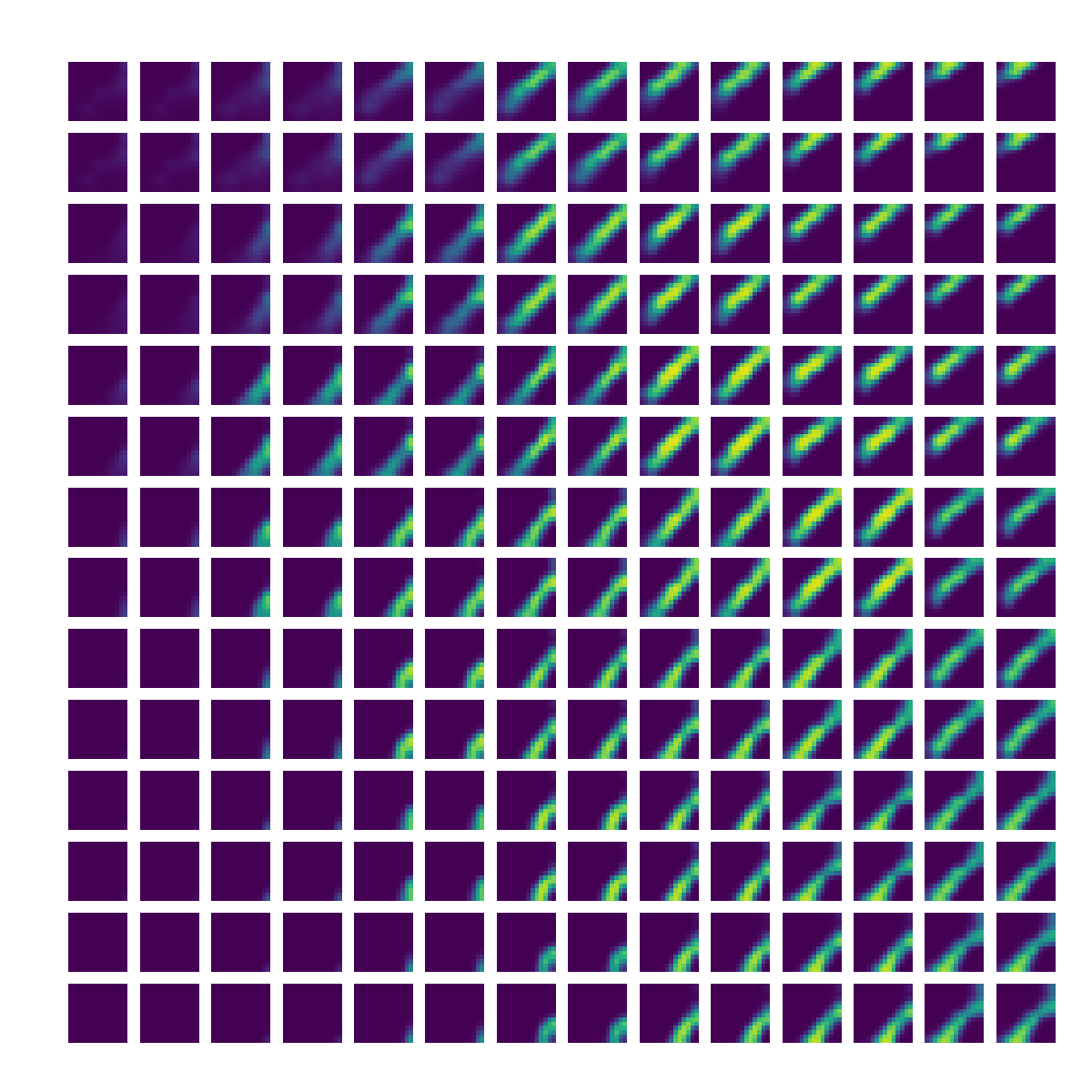}
        \caption{Cross-attention from $\mathbf{\bar I}\rightarrow\mathbf{I}$}
    \end{subfigure}
    \begin{subfigure}{\columnwidth}
        \centering
        \includegraphics[width=0.9\columnwidth]{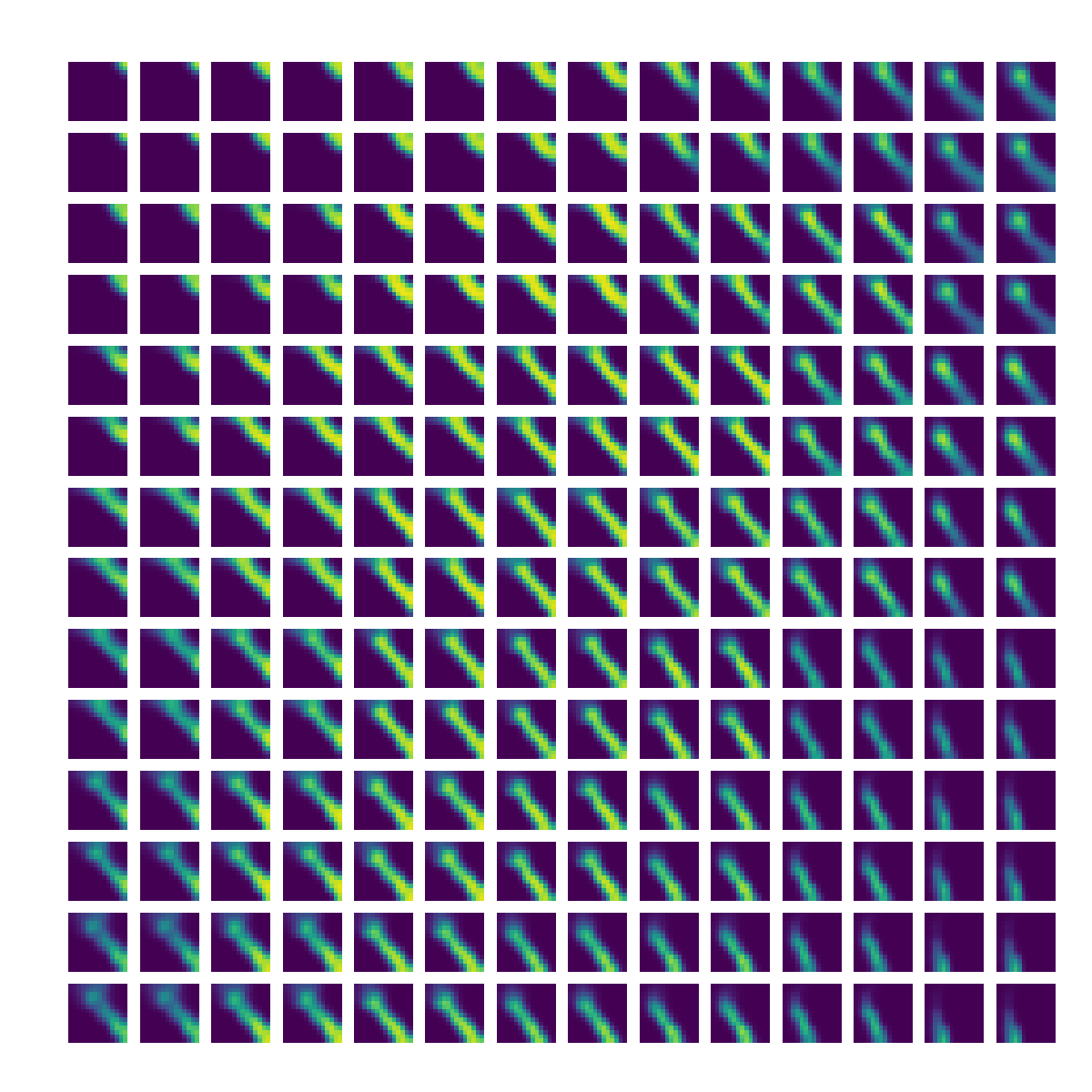}
        \caption{Cross-attention from $\mathbf{I}\rightarrow\mathbf{\bar I}$}
    \end{subfigure}
    \caption{Cross-attention maps extracted from the transformer model trained with $448\!\times\!448$ input images. Due to the higher input resolution, the cross-attention maps are obtained at a higher resolution of $14\!\times\!14\!\times\!14\!\times\!14$.}
    \label{fig:highresmaps}
\end{figure}

In our experiments throughout the paper, the local features tensor obtained from Resnet50 backbone has a spatial resolution of $7\!\times\!7$. We train another transformer model with $L_{EPI}$ on input images of size $448\!\times\!448$ so that we obtain $14\!\times\!14$ local features. By doing this, we can obtain higher resolution ($14\times14\times14\times14$) cross-attention maps, as shown in \cref{fig:highresmaps}. \cref{table:highres-result} shows the performance achieved by the high resolution transformer model.

\begin{table}[t]
\centering
\begin{tabular}{ccccc} 
\toprule 
Model & \makecell{Local features\\ resolution} & $R@1$ & $R@10$ & $R@50$ \\
\midrule
Original & $7\!\times\!7$ &  $90.57$ & $97.33$ & $98.10$ \\
High-res & $14\!\times\!14$ &  $90.71$ & $97.42$ & $98.15$ \\
\bottomrule 
\end{tabular}
\caption{Comparison with transformer model trained on $448\!\times\!448$ images. Both models are trained with $L_{EPI}$. ``Original'' corresponds to the result in \cref{tab:co3d-results}.}
\label{table:highres-result}
\end{table}

\section{Failure Cases}
\begin{figure*}
    \begin{subfigure}{0.45\textwidth}
        \centering
        \includegraphics[width=\textwidth]{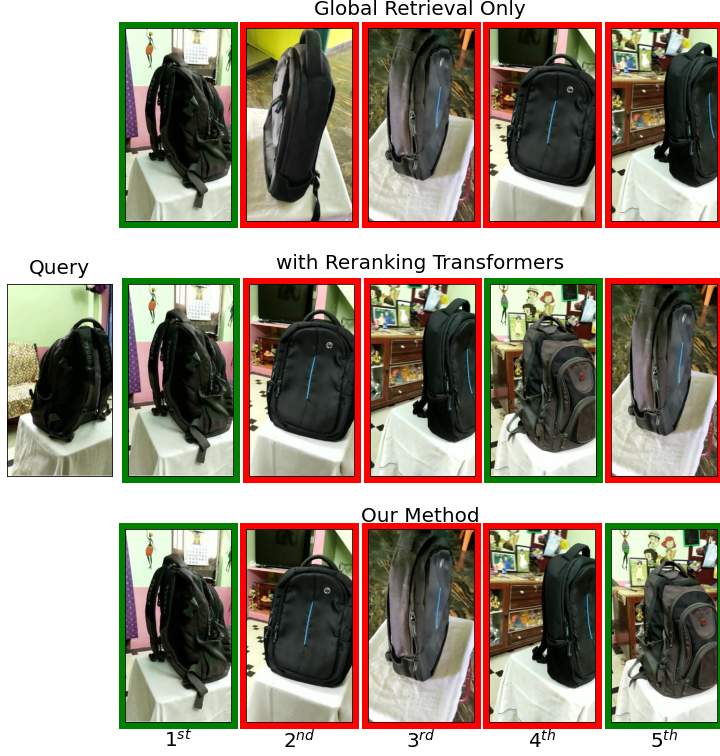}
        \caption{}
    \end{subfigure}%
    \hfill
    \begin{subfigure}{0.45\textwidth}
        \centering
        \includegraphics[width=\textwidth]{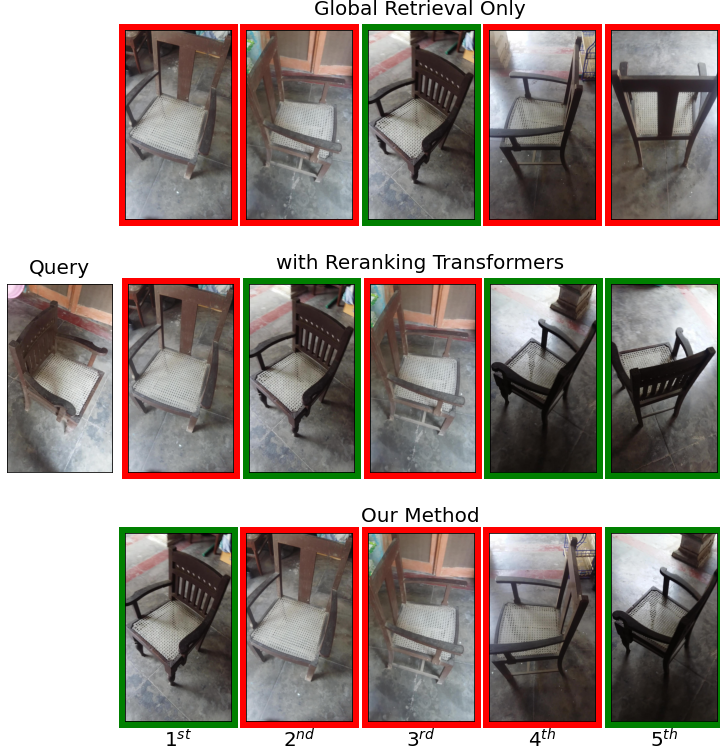}
        \caption{}
    \end{subfigure}
    \\
    \begin{subfigure}{0.45\textwidth}
        \centering
        \includegraphics[width=\textwidth]{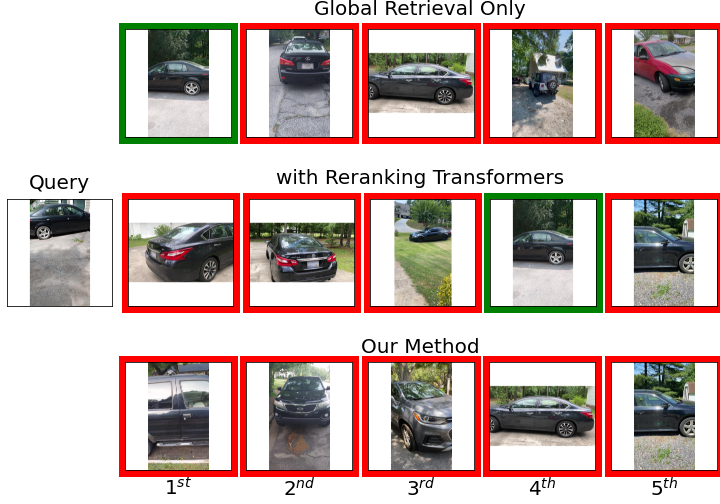}
        \caption{}
    \end{subfigure}%
    \hfill
    \begin{subfigure}{0.45\textwidth}
        \centering
        \includegraphics[width=\textwidth]{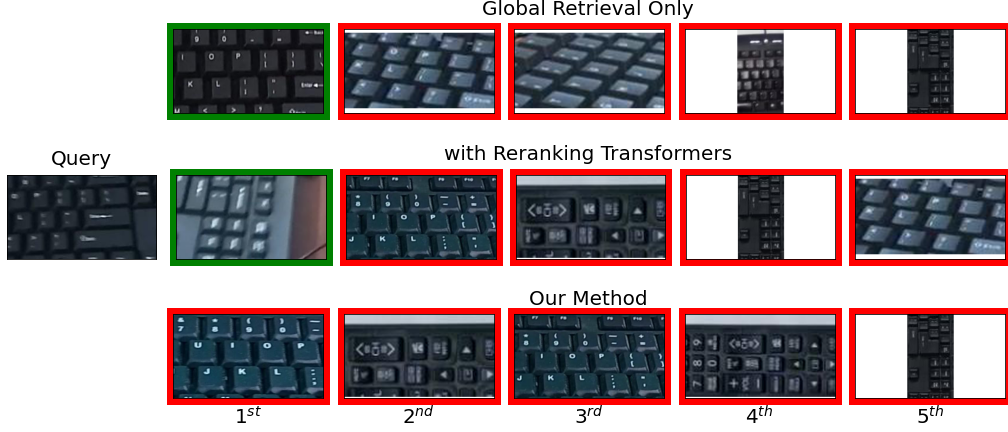}
        \caption{}
    \end{subfigure}
    \caption{Failure cases from the CO3D-Retrieve dataset.}
    \label{fig:fail-co3d}
\end{figure*}

\begin{figure*}
    \begin{subfigure}{0.45\textwidth}
        \centering
        \includegraphics[width=\textwidth]{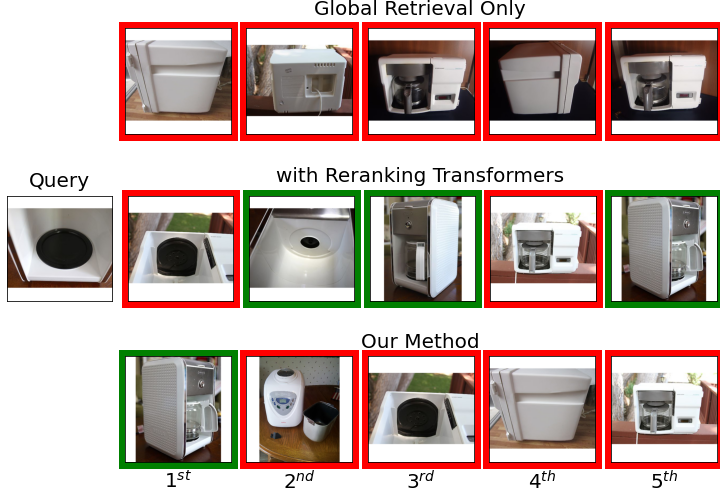}
        \caption{}
    \end{subfigure}%
    \hfill
    \begin{subfigure}{0.45\textwidth}
        \centering
        \includegraphics[width=\textwidth]{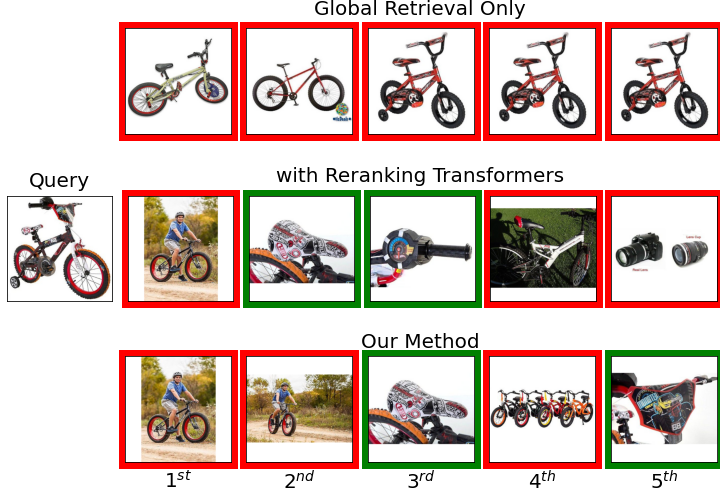}
        \caption{}
    \end{subfigure}
    \\
    \begin{subfigure}{0.45\textwidth}
        \centering
        \includegraphics[width=\textwidth]{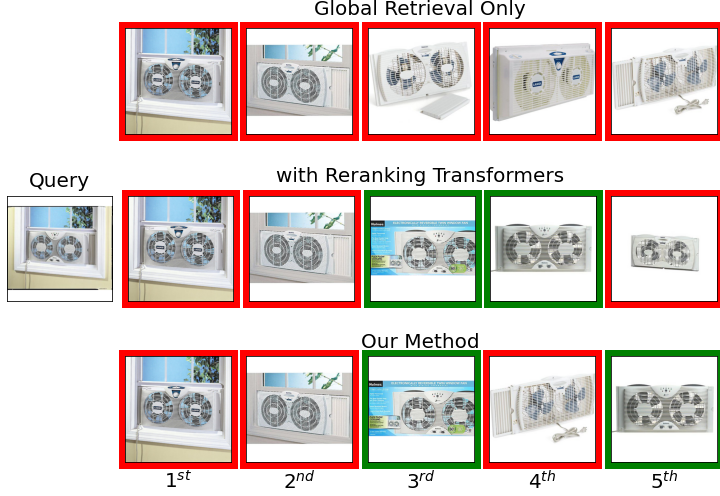}
        \caption{}
    \end{subfigure}%
    \hfill
    \begin{subfigure}{0.45\textwidth}
        \centering
        \includegraphics[width=\textwidth]{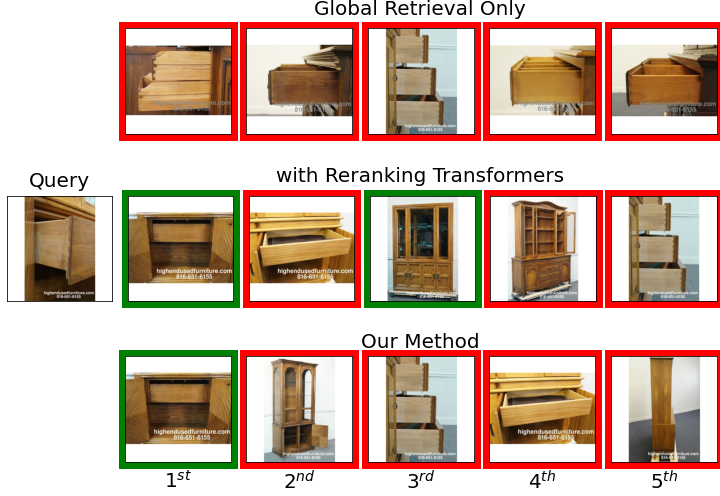}
        \caption{}
    \end{subfigure}
    \caption{Failure cases from the SOP \cite{oh2016deep} dataset.}
    \label{fig:fail-sop}
\end{figure*}

It is important to look at the cases where our proposed method fails to retrieve good matches and analyze them for further improvement. Figures \ref{fig:fail-co3d} and \ref{fig:fail-sop} show a few such examples for CO3D-Retrieve and SOP \cite{oh2016deep} respectively. We see that a common failure scenario for our method is when the query image is a close-up of the object (Fig.\ \ref{fig:fail-co3d} (c,d) and Fig.\ \ref{fig:fail-sop} (c,d)) or repetitive patterns in objects such as keyboards (Fig.\ \ref{fig:fail-co3d} (d)). A critical future direction for our work is to make the model robust to these scenarios.

\section{Quality of Epipolar Geometry with LoFTR/MAGSAC++ method}
\begin{figure*}
    \centering
    \begin{subfigure}{\textwidth}
        \includegraphics[width=\textwidth]{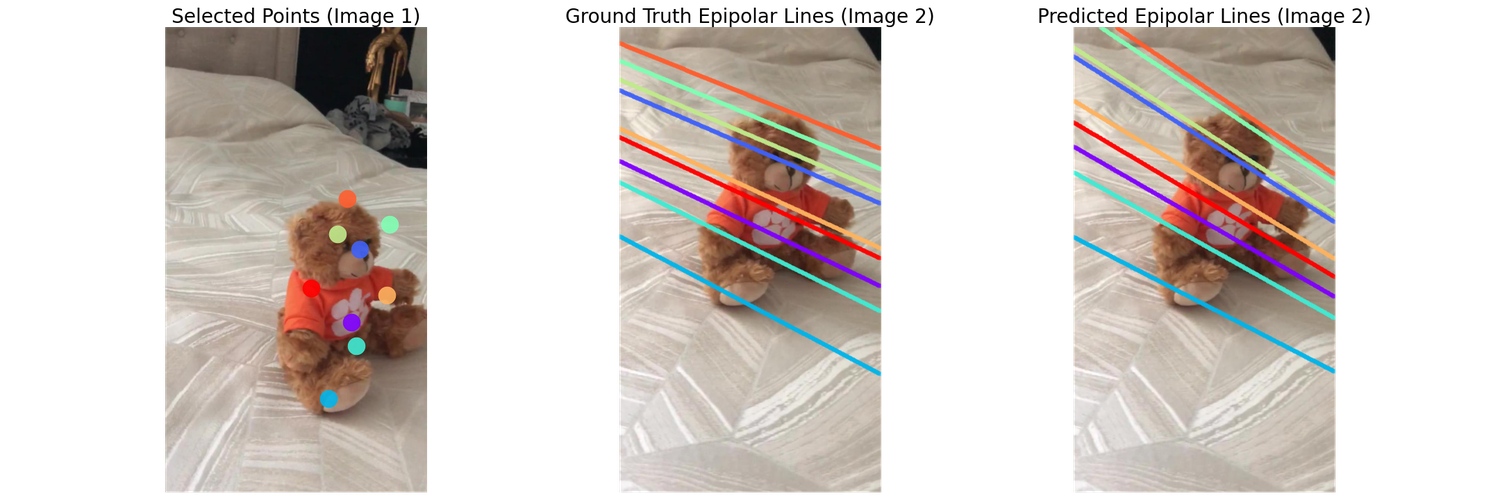}
        \caption{Success case.}
        \label{fig:my_label}  
    \end{subfigure}
    \begin{subfigure}{\textwidth}
        \includegraphics[width=\textwidth]{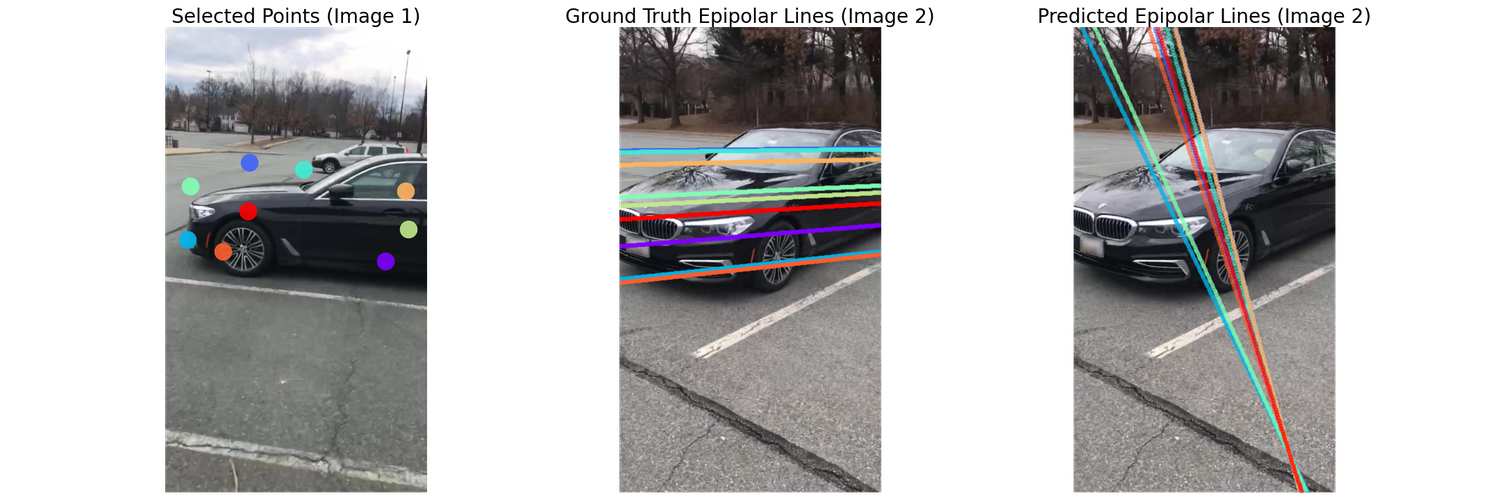}
        \caption{Failure case.}
        \label{fig:my_label}  
    \end{subfigure}
    \caption{Qualitative examples demonstrating the Epipolar geometry predicted using a pretrained LoFTR \cite{sun2021loftr} for matching and MAGSAC++ \cite{barath2020magsac++} for robust optimization.}
    \label{fig:pseudo-geom}
\end{figure*}

During training, when the ground-truth epipolar geometry is not available, we use a \textit{pseudo}-geometry predicted using a pretrained LoFTR \cite{sun2021loftr} model for matching and MAGSAC++ \cite{barath2020magsac++} for robust optimization. The quality of the predicted epipolar geometry depends on the quality and number of matches obtained by the LoFTR model. In Figure \ref{fig:pseudo-geom}, we show two examples demonstrating the success and failure cases of this method.

\begin{figure*}
    \centering
    \resizebox{\textwidth}{!}{%
    \includegraphics[height=6cm]{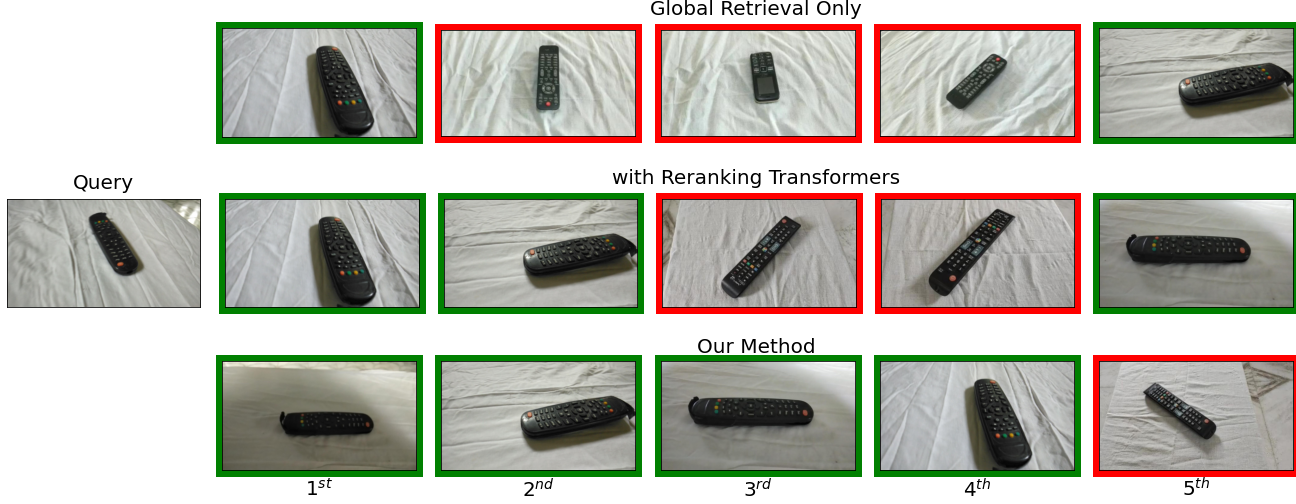}%
    \quad
    \includegraphics[height=6cm]{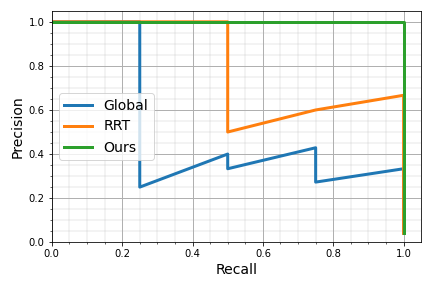}%
    }
    \caption{CO3D-Retrieve: Example 2.}
\end{figure*}

\begin{figure*}
    \centering
    \resizebox{\textwidth}{!}{%
    \includegraphics[height=6cm]{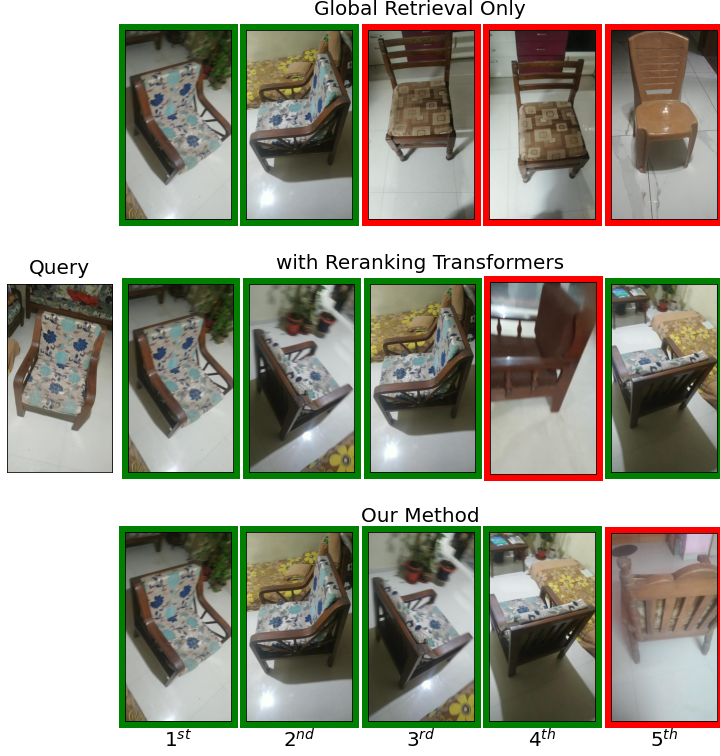}%
    \quad
    \includegraphics[height=4cm]{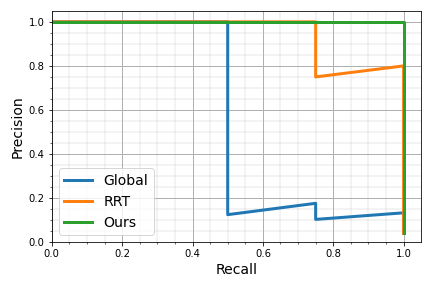}%
    }
    \caption{CO3D-Retrieve: Example 3.}
\end{figure*}

\begin{figure*}
    \centering
    \resizebox{\textwidth}{!}{%
    \includegraphics[height=6cm]{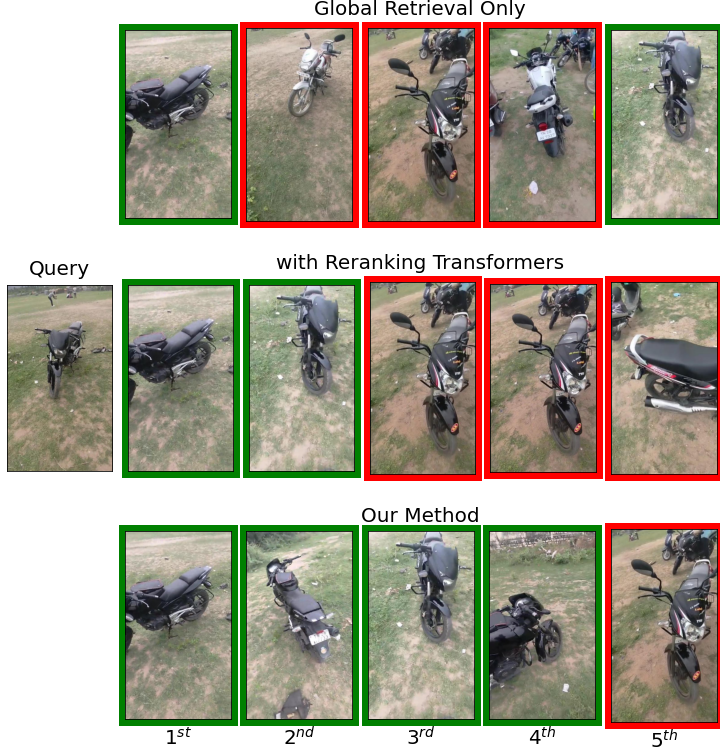}%
    \quad
    \includegraphics[height=4cm]{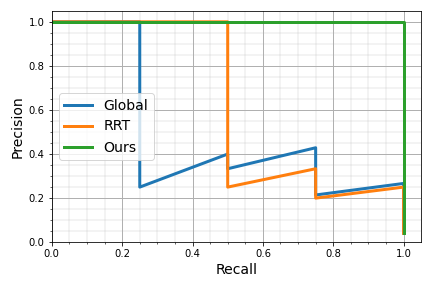}%
    }
    \caption{CO3D-Retrieve: Example 4.}
\end{figure*}

\begin{figure*}
    \centering
    \resizebox{\textwidth}{!}{%
    \includegraphics[height=6cm]{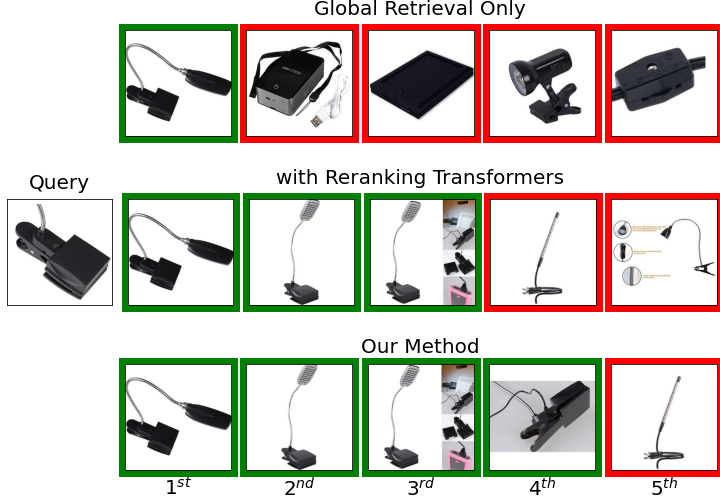}%
    \quad
    \includegraphics[height=5cm]{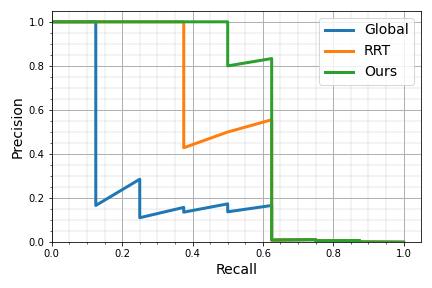}%
    }
    \caption{SOP \cite{oh2016deep} dataset. Example 2.}
\end{figure*}

\begin{figure*}
    \centering
    \resizebox{\textwidth}{!}{%
    \includegraphics[height=6cm]{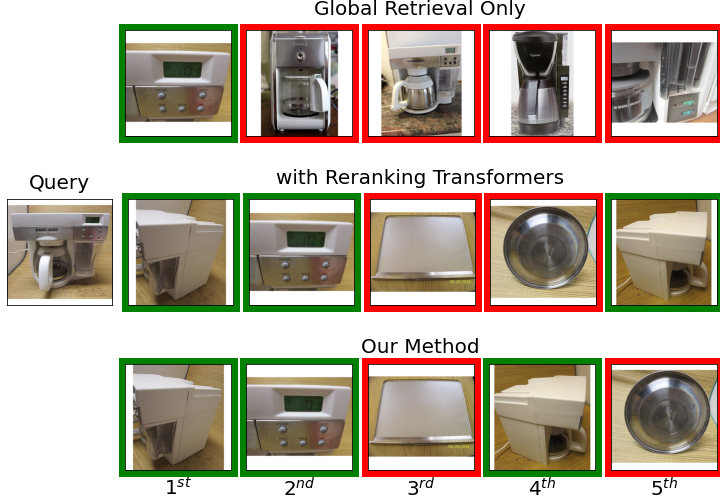}%
    \quad
    \includegraphics[height=5cm]{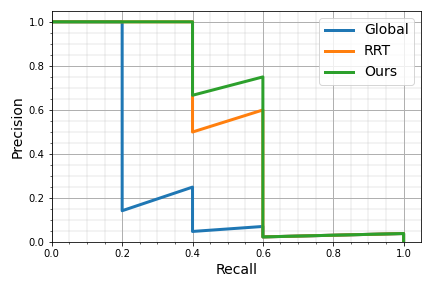}%
    }
    \caption{SOP \cite{oh2016deep} dataset. Example 3.}
\end{figure*}

\begin{figure*}
    \centering
    \resizebox{\textwidth}{!}{%
    \includegraphics[height=6cm]{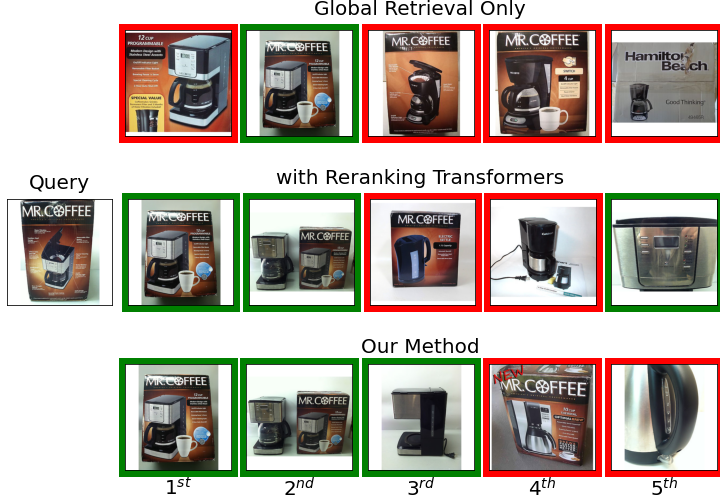}%
    \quad
    \includegraphics[height=5cm]{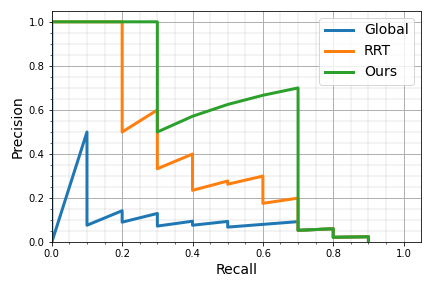}%
    }
    \caption{SOP \cite{oh2016deep} dataset. Example 4.}
\end{figure*}

\begin{figure*}
    \centering
    \resizebox{\textwidth}{!}{%
    \includegraphics[height=6cm]{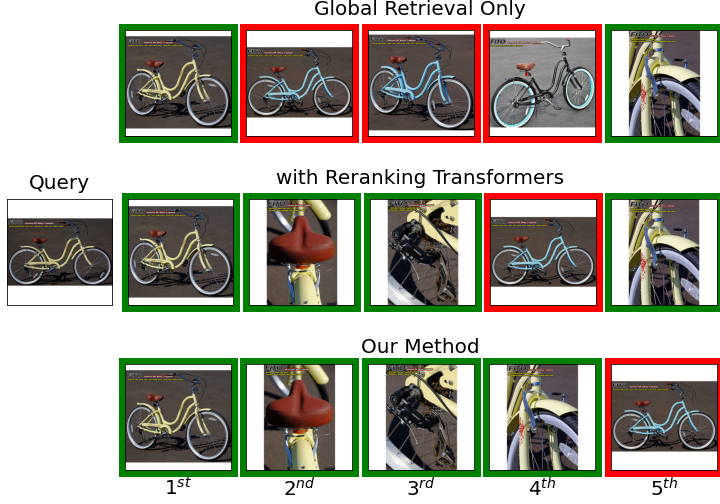}%
    \quad
    \includegraphics[height=5cm]{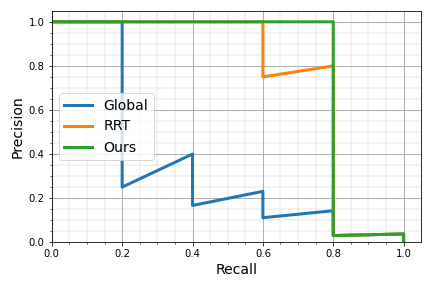}%
    }
    \caption{SOP \cite{oh2016deep} dataset. Example 5.}
\end{figure*}

\end{document}